\documentclass{article}

\usepackage{PRIMEarxiv}

\usepackage[utf8]{inputenc} 
\usepackage[T1]{fontenc}    
\usepackage{hyperref}       
\usepackage{url}            
\usepackage{booktabs}       
\usepackage{amsfonts}       
\usepackage{nicefrac}       
\usepackage{microtype}      
\usepackage{lipsum}
\usepackage{fancyhdr}       
\usepackage{graphicx}       
\graphicspath{{media/}}     
\usepackage[version=4]{mhchem}

\usepackage{microtype}
\usepackage{subcaption}
\usepackage{amsmath,amsfonts,amssymb,amsthm}
\usepackage{mathtools}
\usepackage{commath}
\usepackage{textcomp} 
\usepackage{float} 
\usepackage{amssymb} 
\usepackage{bm} 
\usepackage[ruled,lined,linesnumbered]{algorithm2e}
\usepackage{varwidth}
\usepackage{algpseudocode}
\usepackage{xcolor}

\SetCommentSty{mycommfont}
\usepackage{dblfloatfix}

\usepackage{cleveref}
\crefformat{section}{\S#2#1#3} 
\crefformat{subsection}{\S#2#1#3}
\crefformat{subsubsection}{\S#2#1#3}

\usepackage[final]{changes} 
\definechangesauthor[name=AES, color=blue]{AES}

\newcommand{\beginsupplement}{%
        \setcounter{table}{0}
        \renewcommand{\thetable}{S-\arabic{table}}%
        \setcounter{figure}{0}
        \renewcommand{\thefigure}{S-\arabic{figure}}
        \setcounter{section}{0}
        \renewcommand{\thesection}{S-\arabic{section}}%
        \setcounter{page}{1}
        \renewcommand{\thepage}{S-\arabic{page}}
     }

\title{Fast Bayesian Optimization of Needle-in-a-Haystack Problems using Zooming Memory-Based Initialization (ZoMBI)}

\author{
  Alexander E. Siemenn$^*$\\
  Department of Mechanical Engineering\\
  Massachusetts Institute of Technology\\
  Cambridge, MA 02139, USA\\
  \texttt{asiemenn@mit.edu}\\
   \And
  Zekun Ren\\
  Department of Electrical and Computer Engineering\\
  Singapore-MIT Alliance for Research and Technology\\
  Singapore 138602, Singapore\\
  Xinterra\\
  Singapore 139949, Singapore\\
   \And
  Qianxiao Li\\
  Department of Mathematics\\
  National University of Singapore\\
  Singapore 138602, Singapore\\
  \And
  Tonio Buonassisi\\
  Department of Mechanical Engineering\\
  Massachusetts Institute of Technology\\
  Cambridge, MA 02139, USA\\
}

\begin{document}
\maketitle

\begin{abstract}

Needle-in-a-Haystack problems exist across a wide range of applications including rare disease prediction, ecological resource management, fraud detection, and material property optimization. A Needle-in-a-Haystack problem arises when there is an extreme imbalance of optimum conditions relative to the size of the dataset. For example, only $0.82\%$ out of $146$k total materials in the open-access Materials Project database have a negative Poisson's ratio. However, current state-of-the-art optimization algorithms are not designed with the capabilities to find solutions to these challenging multidimensional Needle-in-a-Haystack problems, resulting in slow convergence to a global optimum or pigeonholing into a local minimum. In this paper, we present a Zooming Memory-Based Initialization algorithm, entitled ZoMBI, that builds on conventional Bayesian optimization principles to quickly and efficiently optimize Needle-in-a-Haystack problems in both less time and fewer experiments by addressing the common convergence and pigeonholing issues. ZoMBI actively extracts knowledge from the previously best-performing evaluated experiments to iteratively zoom in the sampling search bounds towards the global optimum "needle" and then prunes the memory of low-performing historical experiments to accelerate compute times by reducing the algorithm time complexity from $O(n^3)$ to $O(\phi^3)$ for $\phi$ forward experiments per activation, which trends to a constant $O(1)$ over several activations. Additionally, ZoMBI implements two custom adaptive acquisition functions to further guide the sampling of new experiments toward the global optimum. We validate the algorithm's optimization performance on three real-world datasets exhibiting Needle-in-a-Haystack problems that vary in dimensionality from 6D to 11D and further stress-test the algorithm's performance across an additional 174 analytical datasets that vary in optimum needle width, optimum distance to edges, dimensionality, and initialization conditions. The ZoMBI algorithm demonstrates compute time speed-ups of 400x compared to traditional Bayesian optimization as well as efficiently discovering optima in under 100 experiments that are up to 3x more highly optimized than those discovered by similar methods MiP-EGO, TuRBO, and HEBO.

\end{abstract}

\keywords{rare materials discovery \and efficient algorithms \and adaptive acquisition functions \and trust regions \and optimization \and extremely imbalance data \and auxetic materials \and thermoelectric materials, \and active learning}

\section{Introduction}

\begin{figure}[h!]
\centering
\begin{subfigure}[b]{0.43\textwidth} 
\includegraphics[width=\textwidth]{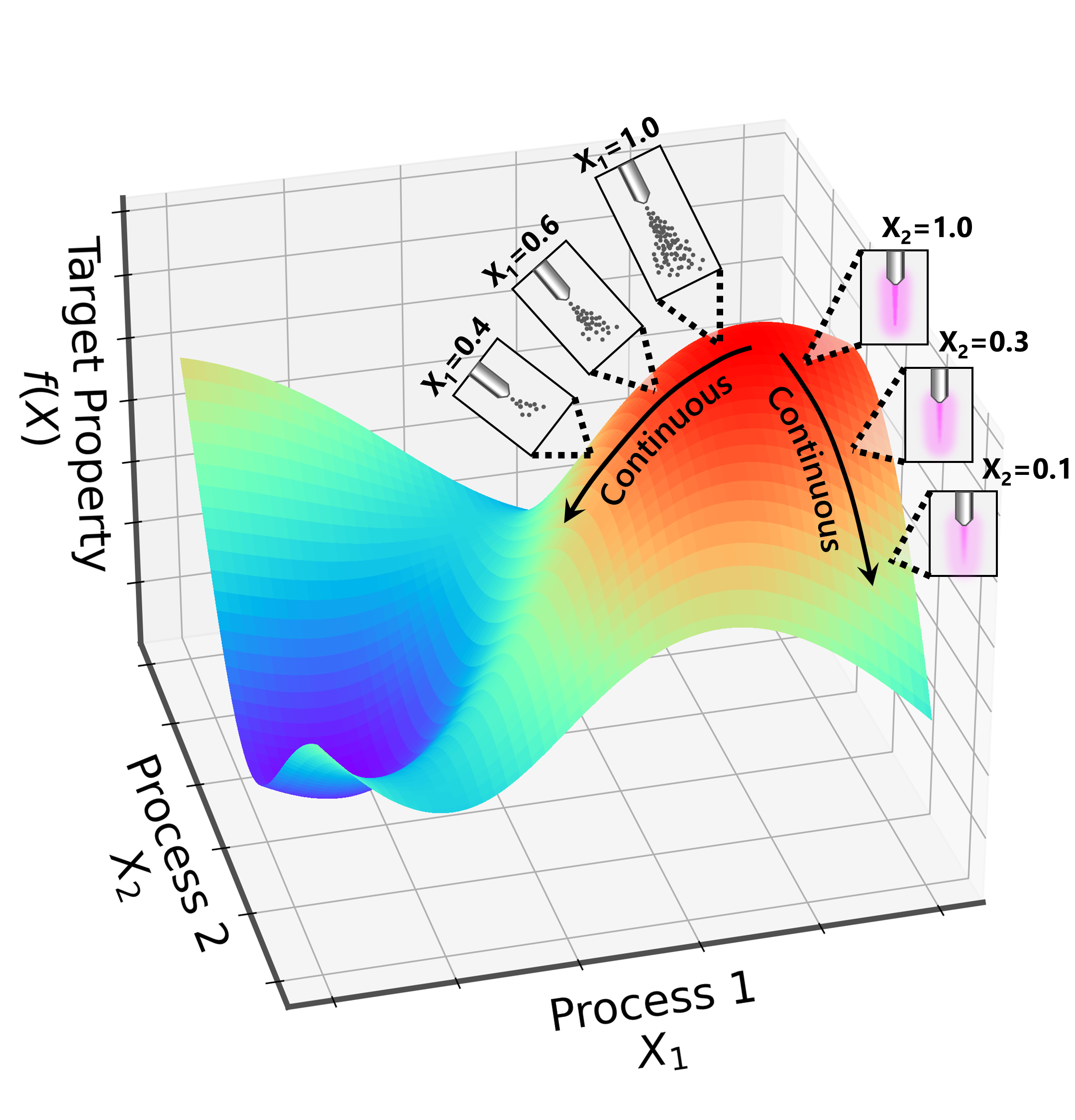}
\caption{Process Optimization Manifold}
\end{subfigure}\hfill%
\begin{subfigure}[b]{0.43\textwidth} 
\includegraphics[width=\textwidth]{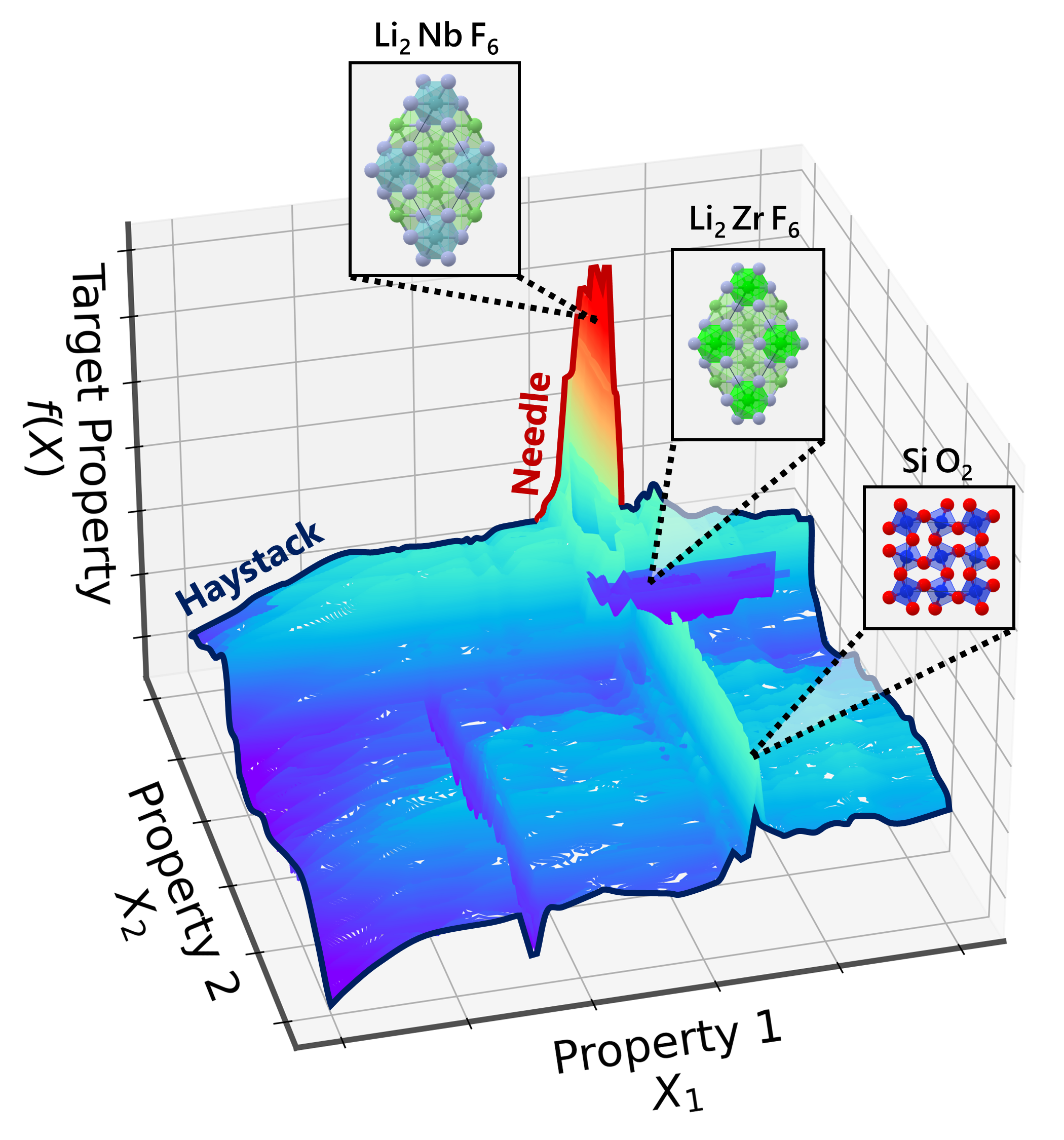}
\caption{Materials Optimization Manifold}
\end{subfigure}\hfill%
\caption{\textbf{Archetypal Manifolds in Materials Science Optimization.} (a) In process optimization, there often exists a real and continuous path between each condition. This 3D projected manifold is adapted from the 6D perovskite process optimization problem by Liu \textit{et al.}, where $X_1$ is spray flow rate, $X_2$ is plasma voltage, and $f(X)$ is cell efficiency \cite{Liu2022}. (b) However, in materials optimization, there are often only discrete combinations of properties that define real materials, resulting in a rough topology with extreme outliers. For example, Li$_2$NbF$_6$ and Li$_2$ZrF$_6$ lay close to each other in space because they have similar density, formation energy, and structure, however, they have vastly different target properties: Li$_2$NbF$_6$ has a Poisson's ratio of $-1.7$ while Li$_2$ZrF$_6$ has a Poisson's ratio of $0.3$ \cite{DeJong2015}. Extreme outliers, such as Li$_2$NbF$_6$, consist of only a small fraction of the manifold hypervolume, resulting in a Needle-in-a-Haystack regime arising. This 3D projected manifold is obtained from the 6D Poisson's ratio optimization problem presented in this paper, where $X_1$ is density, $X_2$ is formation energy, and $f(X)$ is negative Poisson's ratio \cite{Jain2013}.}
\label{fig:manifold}
\end{figure}

Current optimization algorithms achieve good results on low-dimensional problems that are smooth and have wide basins of attraction. Examples of smooth manifolds with wide basins of attraction within material science include process- and recipe-optimization problems such as tuning perovskite manufacturing variables to achieve higher efficiency \cite{Liu2022}, optimizing microfluidics flow parameters to achieve ideal droplet formation \cite{Siemenn2022}, optimizing silver nanoparticle recipes for optical properties \cite{Mekki-Berrada2021}, and tuning perovskite compositions with physics-based constraints to maximize stability \cite{Sun2021}. Optimization techniques like Bayesian optimization (BO) are well-suited to model these simple manifolds using a Gaussian Process (GP) surrogate \cite{snelson2005, Rasmussen2005, Brochu2010, Snoek2001, Liang2021}. However, the performance of this BO with a GP breaks down as the manifold complexity increases. Material property optimization problems that have high technological significance, such as discovering materials with rare properties or materials with a specific combination of properties, have search space manifolds that more closely resemble a \textit{Needle-in-a-Haystack} \cite{Kim2020}, shown in Figure \ref{fig:manifold}(b), rather than a smooth or convex space.

This Needle-in-a-Haystack (NiaH) problem arises when only few optimum conditions exist within the entire dataset, resulting in an extreme imbalance. Interpolating the parameter space of an imbalanced dataset with an estimation function, such as a GP, results in smoothing over the optimum or over-predicting the properties of the materials found near the optimum \cite{Andricioaei1996, Seeger2004, Snoek2015}. Examples of NiaH materials optimization problems include discovering auxetic materials (\textit{i.e.}, materials that have a highly negative Poisson's ratio, $\nu$) for energy absorptive medical devices or protective armor \cite{Dagdelen2017,Saxena2016, Liu2006} and discovering materials that have a combination of high electrical conductivity and low thermal conductivity (\textit{i.e.}, a highly positive thermoelectric figure of merit, $ZT$) used for improving sensor technology to enable ubiquitous solid-state cooling \cite{Salah2020, He2018, Mao2020}. Optimization of these rare material properties illustrates examples where an extreme data balance exists in the dataset because only a fraction of the total number of materials exhibit these rare properties \cite{Dagdelen2017, Jain2013, DeJong2015, Yeganeh-Haeri1992, Lakes2008}. This NiaH optimization challenge of extremely imbalanced datasets is largely applicable to many fields, not just materials science, including the fields of ecological resource management \cite{Rew2006, Bouguettaya2022}, fraud detection \cite{Wei2012,Marchant2021}, and rare diseases \cite{Khalilia2011, Marchant2021}.

Several challenges exist for the current landscape of computational tools that inhibit effective optimization of these complex NiaH problems. Firstly, the "needle" makes up only a small percentage of the total manifold search space, resulting in a weak correlation between the measured input parameters and the target property of interest, inhibiting discovery of the region containing the needle \cite{Crammer2004,Liu2018,Andricioaei1996}. This challenge requires the development of an algorithm that can more quickly determine the plausible region of the manifold where the needle exists. The second challenge for algorithms, such as BO, to optimize NiaH manifolds is in the nature of the acquisition function to pigeonhole sampling into local minima because of the narrowness of the needle's basin of attraction \cite{Nusse1996, Datseris2022}. Standard BO acquisition functions, including expected improvement (EI) \cite{Hennig2012} and lower confidence bound (LCB) \cite{Brochu2010, Seeger2004}, are static sampling techniques that only adjust sampling based on the output of the surrogate model, which enacts smoothing of the needle \cite{Andricioaei1996, snelson2005, Rasmussen2005}. To overcome this challenge, active learning-based tuning of the acquisition function hyperparameters can be implemented to improve the sampling quality and avoid pigeonholing. Lastly, there exists a computing challenge for NiaH problems where, typically, several thousands of samples must be observed to find an optimum when using an algorithm that is poorly suited to tackle NiaH manifolds \cite{Kim2020}. The compute time of BO using a GP surrogate scales with the complexity $O(n^3)$, where $n$ is the number of experiments sampled, hence, the compute time of traditional BO blows up as more data is required to find the optimum \cite{BelyaevMikhail2014,Li2017, Wang2017, snelson2005, Rasmussen2005, Bui2017, Lan2020}. To solve this computing challenge, an algorithm must be designed that both efficiently optimizes the space in as few experiments as possible and reduces the effect of compounding compute times over the length of the optimization procedure.

In recent literature, algorithms have been developed to address some of these challenges individually, but not all of them together. The first class of solutions bound the search space using a trust region approach to sample regions with higher probability of containing the optimum. Eriksson \textit{et al.} develop \texttt{TuRBO} \cite{Eriksson2020} that compiles a set of independent model runs, using separate GP surrogate models to compute a new, smaller search region, narrowed in on the target optimum. Regis develops \texttt{TRIKE} \cite{Regis2015} that utilizes maximization of the EI acquisition function to bound a trust region containing the global optimum. Diouane \textit{et al.} develop \texttt{TREGO} \cite{Diouane2021}, which interleaves sampling between global and local search regions, where the local search regions are defined by the single best historical experiment sampled. Although these methods offer solutions to one of the three challenges presented, each method has its downfalls when optimizing NiaH problems. For example, \texttt{TuRBO} requires the computation of several GP model runs, which increases compute time and also does not guarantee that the needle will be resolved due to interpolation effects; \texttt{TRIKE} is inflexible to the use of other acquisition functions as it locks the user in to only using EI, which may pigeonhole into local minima; \texttt{TREGO} uses only the best sampled experiment to define its search regions, which will yield inconsistent or sub-optimal results when the needle consists of a fractional region of the manifold and single point is unlikely to land in its basin of attraction. The second class of solutions to the challenges presented in this paper are designed to decrease the computing time required to run an optimization procedure. A common method for reducing the compute time of BO with a GP surrogate is to introduce a sparse GP \cite{snelson2005, titsias09a, Bui2017}. A sparse GP uses a small subset of pseudo data, often denoted as $m$, to reduce the GP time complexity from $O(n^3)$ to $O(nm^2)$ \cite{Leibfried2021}. However, the process of selecting a useful subset requires minimizing the Kullback-Leibler divergence between the sparse GP and true posterior GP, which is often a computationally intensive procedure of using variational inference \cite{turnersahani2011}. In addition to sparse GPs, new algorithms have been developed in literature to improve the compute time of optimization in various ways. Van Stein \textit{et al.} develop \texttt{MiP-EGO} \cite{van2019automatic}, which parallelizes the function evaluations of efficient global optimization (EGO) to discover optima faster and in fewer experiments using derivative-free computation \cite{Jones1998}. Joy \textit{et al.} \cite{Joy2020} use directional derivatives to accelerate hyperparameter tuning by 100x and achieve higher accuracy than the \texttt{FABOLAS} baseline by Klein \textit{et al.} \cite{Klein2017}. Zhang \textit{et al.} develop \texttt{FLASH} \cite{Zhang} to achieve optimization speed-ups of 50\% by using a linear parametric model to guide algorithm search within high-dimensional spaces. Snoek \textit{et al.} \cite{Snoek2015} design a neural network-based parametric model that reduces the overall time complexity of BO to $O(n)$ compared to the complexity of $O(n^3)$ of standard BO with a GP surrogate model. These existing methods from literature within the class of solutions for accelerating compute time are generally introducing external models necessary to perform optimization, such as neural networks, variational inference, or parametric models. While these external models do speed up compute time, they often lack the predictive capabilities to capture the weak correlation between measured input parameters and the target property of interest in NiaH problems. We illustrate this mechanism later in the paper when comparing the optimization results on two materials science NiaH problems of a fast algorithm \texttt{MiP-EGO} with that of \texttt{TuRBO}, an algorithm better suited for discovering optima within narrow basins of attraction.



\begin{figure}[h!]
\centering
\begin{subfigure}[b]{1\textwidth}  
\includegraphics[width=\textwidth]{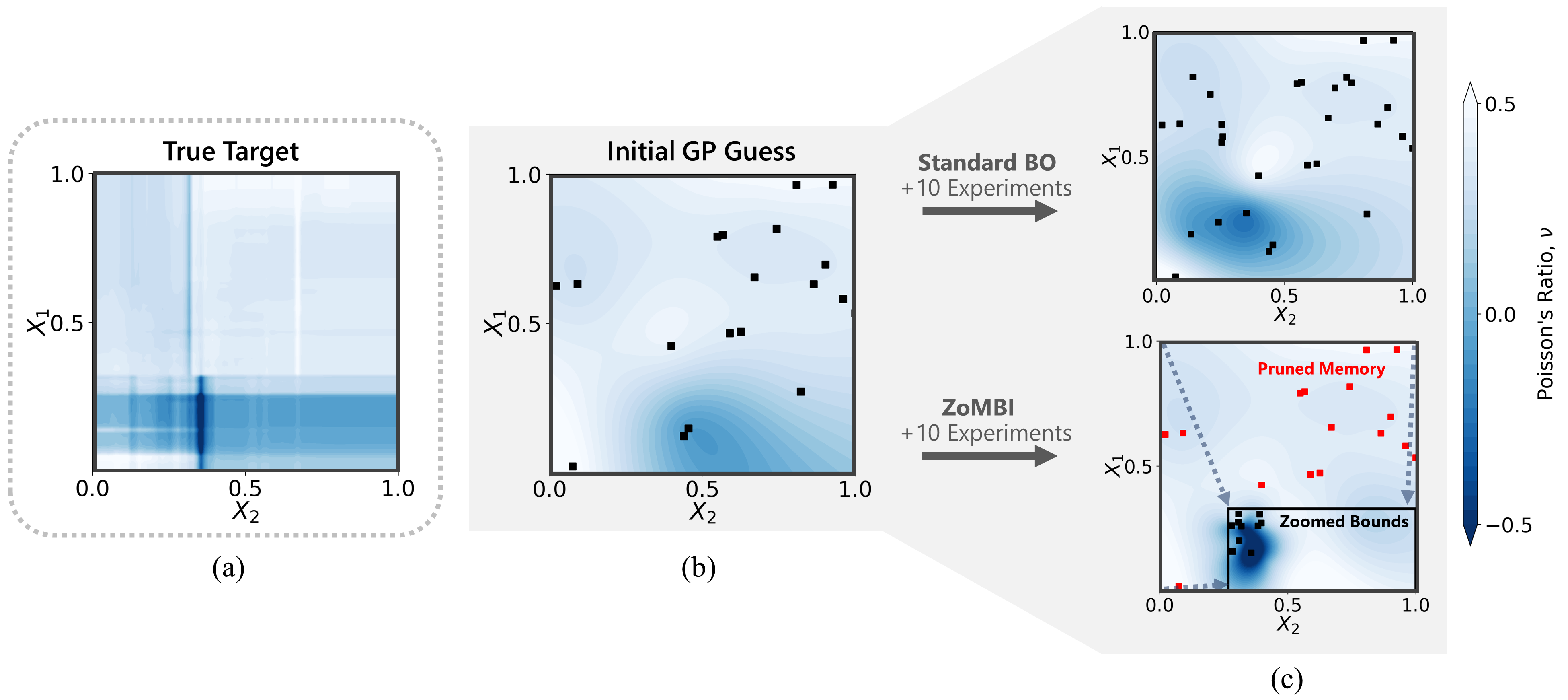}
\end{subfigure}\hfill%
\caption{\textbf{Accelerated Convergence to True Target using \texttt{ZoMBI}.} Using a standard Bayesian optimization procedure, the discovery of a Needle-in-a-Haystack condition does not progress significantly after 10 additional experiments from the initial GP guess. However, using \texttt{ZoMBI} to zoom the bounds inward and prune redundant memory points, the needle-like optimum region is resolved to be accurately aligned with the true target. (a) The true target to optimize, which is a slice from the 6D Poisson's Ratio dataset, (b) The initial guess of the target function using a GP surrogate with 20 randomly sampled experiments, (c) (top) The estimated target resolved by standard BO after 10 additional experiments sampled using a greedy LCB acquisition function ($\beta=0.1$); (bottom) the estimated target resolved by \texttt{ZoMBI} after 10 additional experiments sampled using the same greedy LCB acquisition function. The red memory points do not assist in resolving this target after zooming in the bounds, hence, they are pruned from memory by \texttt{ZoMBI}.}
\label{fig:prune}
\end{figure}

Although these methods from existing literature address some of the challenges in optimizing NiaH problems, none of them have been designed specifically to quickly and efficiently discover a needle-like optimum within a haystack of sub-optimal points, resulting in all of them falling short of a full solution. Therefore, in this paper, we design an algorithm that addresses all three of the challenges faced when optimizing NiaH problems by (1) zooming in the manifold search bounds iteratively and independently for each dimension based on $m$ number of best memory points to quickly converge to the plausible region containing the global optimum needle, (2) relieving compute utilization by pruning the low-performing and redundant memory points not being used to zoom in the search bounds, (3) anti-pigeonholing into local minima by using actively learned acquisition function hyperparameters to tune the exploitation-to-exploration ratio. The proposed algorithm, entitled [Zo]oming [M]emory-[B]ased [I]nitialization (\texttt{ZoMBI}), combines these three contributions into a method that efficiently optimizes NiaH problems quickly. Figure \ref{fig:prune} demonstrates the accelerated convergence ability of the proposed (\texttt{ZoMBI}) algorithm compared to standard BO. In essence, this process of scanning broadly and then focusing in on points of interest based on memory was inspired by the way we humans solve similar problems, but stands in contrast to the way standard BO methods with static acquisition functions solve problems. We demonstrate the performance of this algorithm on three vastly different NiaH problems in materials science and ecological resource management: (1) discovery of materials with negative Poisson's ratio, (2) discovery of materials with both high electrical conductivity and low thermal conductivity, and (3) detection of environmental conditions conducive of sustaining wildfires. The performance of the proposed \texttt{ZoMBI} algorithm is compared against standard BO with static acquisition functions as well as against three more algorithms: (1) \texttt{HEBO}, the winning submission of the NeurIPS 2020 Black-Box optimization challenge \cite{hebo} and one algorithm from each of the two classes of partial NiaH solutions (2) \texttt{TuRBO} (bounded search space) \cite{Eriksson2020} and (3) \texttt{MiP-EGO} (faster compute) \cite{van2019automatic}. Finally, we stress-test the proposed \texttt{ZoMBI} algorithm across 174 additional datasets varying the optimum needle width, optimum distance to edges, dimensionality, and initialization conditions.

\section{Methodology}

In this paper, we develop two major contributions: (1) the \texttt{ZoMBI} algorithm and (2) adaptive learning acquisition functions. Through the combination of these two contributions, the optimum region of a NiaH manifold can be quickly discovered in fewer experiments without pigeonholing into local minima. Thus, the three challenges of optimizing NiaH problems are addressed: (1) the challenge of finding a hypervolume within the manifold that contains the needle-like optimum \cite{Crammer2004,Liu2018,Andricioaei1996}, (2) the challenge of the polynomially increasing compute times of BO using a GP surrogate \cite{Li2017, Wang2017, snelson2005, Rasmussen2005, Bui2017, Lan2020}, (3) the challenge of avoiding pigeonholing into local minima \cite{Liang2021, Liu2022,Nusse1996, Datseris2022}. We demonstrate the implementation of \texttt{ZoMBI} on a 6D analytical Ackley function, a 6D dataset of materials with Poisson's ratios, a 6D dataset of thermoelectric materials, and an 11D dataset for wildfire detection, all of which exhibit an extreme data imbalance and a NiaH regime, and compare the performance to that of \texttt{MiP-EGO} \cite{van2019automatic}, \texttt{TuRBO} \cite{Eriksson2020}, and \texttt{HEBO} \cite{hebo}. For each of the three problems, the objective is to find the target value, $y$, with either the lowest or highest value depending on if the problem is minimization or maximization. This optimum $y$-value resembles a needle for each problem because it is located within a narrow and steep basin of attraction. Precisely, the needle optimum for each problem has a value of $y=0$ for the Ackley function (minimization), $y=-1.7$ for Poisson's ratio dataset (minimization), $y=1.9$ for the thermoelectric merit dataset (maximization), and $y=-12$ for the wildfire detection dataset (minimization). To extend the applicability of \texttt{ZoMBI} optimization performance results to a wider array of applications, additional stress tests are conducted on 174 analytical datasets. First, a set of 144 analytical datasets are optimized to assess the failure and success conditions of \texttt{ZoMBI} on problems with extremely narrow optima and few initialization data points. Then, in the Supplemental Information, a set of 30 analytical datasets are optimized to assess the failure and success conditions of \texttt{ZoMBI} on problems with insufficient initialization data and cases where the global optimum is near the edge of the manifold. 

\subsection{Zooming Memory-Based Initialization (\texttt{ZoMBI}) Algorithm}
\label{sec:zombi}

The \texttt{ZoMBI} algorithm has two key features: (1) iterative inward bounding of proceeding search spaces using the $m$ number of best-performing memory points within the prior search space and (2) iterative pruning of low-performing historical search space memory. The newly computed search space bounds are unique for each dimension, such that the optimum basin of attraction of complex, non-convex NiaH manifolds can be discovered. This algorithm leverages these two key features to guide the acquisition of new data towards more optimal regions while only fitting the surrogate within the suggested optimum region to resolve more detail of the space of interest, as shown in Figure \ref{fig:zombi} and Figure \ref{fig:prune}. This process subsequently reduces the compute time significantly compared to the compute of a GP in a standard BO procedure, as shown in Figure \ref{fig:compute}.

\IncMargin{2em}
\begin{algorithm}[h!]
\DontPrintSemicolon
   \caption{Zooming Memory-Based Initialization (\texttt{ZoMBI})}
   \label{alg:zombi}
  \SetKwData{Left}{left}
  \SetKwData{Up}{up}
  \SetKwFunction{FindCompress}{FindCompress}
  \SetKwInOut{Input}{Input}
  \SetKwInOut{Output}{Output}
  \SetKwRepeat{Repeat}{repeat}{end}

\Indm\Indmm
    \BlankLine
  \Input{\qquad$\mathbf{X}$: Set of data points $\{X_1,X_2,\dots,X_n\}$, where $X_j \in \mathbb{R}^d$,\\
  \qquad$\mathbf{y}$: Set of target values $\{y_1,y_2,\dots,y_n\}$, where $y_j \in \mathbb{R}$,\\
  \qquad $\alpha$: Number of \texttt{ZoMBI} activations,\\
  \qquad $\phi$: Number of forward experiments per activation,\\
  \qquad $\bm{\gamma}$: Set of acquisition function hyperparameters $\{\beta, \xi, \epsilon, \eta \}$,\\
  \qquad $AF$: An acquisition function selected by the user}
  \BlankLine
  \Output{\qquad The next experimental condition $X_{n+1} \in \mathbb{R}^d$ and measured target value $y_{n+1}\in \mathbb{R}$}
\Indp\Indpp
  \BlankLine
    \For{$\alpha$ \textup{activations}}{
    Compute bounds $\{\mathcal{B}^l_d,\mathcal{B}^u_d\} \leftarrow \{\min,\max\}_{X \in \mathbf{X}^{(m)}} \{X_d\}$\\
    \begin{varwidth}[t]{\linewidth}
    Initialize with $i$ LHS data points $\{X\}:=\{X_1,X_2,\dots,X_i\}$, where $X_j\in\mathbb{R}^d,[\mathcal{B}^l_d,\mathcal{B}^u_d]$\par
    \hskip\algorithmicindent and target values $\{y\}:=\{y_1,y_2,\dots,y_i\}$, where $y_j\in\mathbb{R}$ 
    \end{varwidth}\\
    Overwrite memory $\mathbf{X} \leftarrow \{X\}$ and $\mathbf{y} \leftarrow \{y\}$\\ 
    \For{$f$ \textup{in} $\texttt{range}(1,\phi)$ \textup{forward experiments}}{
    Let $n = i+f$\\
    Retrain surrogate model $\mathcal{GP}(\mathbf{X})$ using target values $\mathbf{y}$\\ 
    Extract set of surrogate means $\bm{\mu}$ and variances $\bm{\sigma}$\\  
    Compute set of acquisition values $\bm{a} \leftarrow AF(\bm{\mu}, \bm{\sigma}; \bm{\gamma})$\\ 
    Find the best new experimental condition $X_{n+1}\leftarrow$ \textrm{arg\,max}$\left(\bm{a}\right)$\\ 
    Measure target value of new experimental condition $y_{n+1}$\\ 
    Append outputs to sets $\mathbf{X}\texttt{.append}(X_{n+1})$ and $\mathbf{y}\texttt{.append}(y_{n+1})$}
    }
\end{algorithm}

We define $m$ as the number of retained memory points during an activation of \texttt{ZoMBI}. The $m$ memory points are saved to memory while all other data are erased from memory. These are the historical data points that achieve the $m$ lowest (for minimization) target values, $y$, and they are used to zoom in the search bounds. Using these memory points, the multi-dimensional upper and lower bounds of the zoomed search space are computed for each dimension, $d$. Let $\mathbf{X}:=\{X_1,X_2,\dots,X_n\}$ be a set of data points, where $X_j \in \mathbb{R}^d$. Let $f:\mathbb{R}^d \rightarrow \mathbb{R}$ be the objective function. We first assume that the points in $\mathbf{X}$ are in general position so that $f(\mathbf{X})$ contains unique elements. Then, for each $m \leq n$ define $\mathbf{X}^{(m)}=\{ X_{\pi(1)},\dots,X_{\pi(m)} \}$ where $\pi$ is a permutation on $\{1,\dots,n\}$ so that $\{f(X_{\pi(j)})\}$ is in ascending order. If $f(\mathbf{X})$ contains repeated elements, we may first remove the points with repeated $f$ values and apply the definition above. Then, for each $d$, the bounds are defined as:
\begin{align}
\begin{split}
\label{eq:bounds}
    \mathcal{B}^l_d &= \min_{X \in \mathbf{X}^{(m)}} \{X_d\}\\
    \mathcal{B}^u_d &= \max_{X \in \mathbf{X}^{(m)}} \{X_d\},
\end{split}
\end{align}
where $\mathcal{B}^l_d$ and $\mathcal{B}^u_d$ computed lower and lower bounds for each dimension, $d$, respectively. The bounds $[\mathcal{B}^l_d, \mathcal{B}^u_d]$ constrain the proceeding acquisition of new data as well as the computation of a GP, such that sampling cannot occur outsides of the bounded region. This constraining process operates independently for each dimension, such that each dimension has a unique lower and upper bound. To initialize the algorithm with data from the constrained space, $i$ data points are sampled from the bounded region using Latin Hypercube Sampling (LHS). LHS splits a $d$-dimensional space into $i*d$ equally spaced strata, where $i$ is the number of points to sample uniformly over $d$ dimensions with low variability, unlike random sampling that has high sampling variability \cite{McKay2000}. A GP surrogate model is retrained on these $i$ LHS points sampled from the constrained space and then for every proceeding experiment sampled from the space, denoted as a forward experiment, the surrogate model is retrained. Thus, the GP is only being trained on information within the constrained region and as the constrained region iteratively zooms inward and decreases in hypervolume, so does the region computed by the GP. This process allows for more information to be resolve within regions plausibly containing the global optimum basin of attraction. Up to $\phi$ forward experiments are sampled in serial, where $\{X_i\} \cup \{X_\phi\} \subseteq \{X_n\}$. These forward experiments are sampled by maximizing an acquisition value, $a\in[0,1]$, computed by a user-selected acquisition function from one of the four functions EI, EI Abrupt, LCB, and LCB Adaptive, introduced in Section \ref{sec:acquisition}. Once $i+\phi$ number of experiments are sampled, the bounds are re-constrained using the $m$ best performing experiments, $i$ new experiments are sampled from the zoomed-in space using LHS, and then the memory is pruned. The process of collecting $\phi$ forward experiments is repeated. A complete constraining-resetting iteration is denoted as an activation, $\alpha$. This iterative zooming and pruning process over several $\alpha$ significantly speeds up compute time, discussed further in Section \ref{sec:compute}. Implementation of \texttt{ZoMBI} is shown in Algorithm \ref{alg:zombi}.

\subsection{Adapative Acquisition Functions}
\label{sec:acquisition}

Traditional BO acquisition functions, such as EI \cite{ei} and LCB \cite{Auer2002}, use the computed means and variances from a surrogate model to compute an acquisition value; maximizing this acquisition value guides sampling of the manifold \cite{Brochu2010, Hennig2012, Seeger2004}. However, these traditional acquisition functions are static, such that they do not actively use any information about the performance of previously sampled experiments to guide sampling. Hence, we implement an adaptive learning approach into the acquisition functions to develop two novel functions, EI Abrupt and LCB Adaptive, that dynamically adapt their sampling based on the quantity and quality of previously sampled experiments. In contrast to a static acquisition function, these adaptive acquisition functions are initialized with an initial set of hyperparameter values to guide their search but then tune these values as sampling progresses. The developed EI Abrupt and LCB Adaptive functions are used within the \texttt{ZoMBI} framework to further accelerate optimization and avoid pigeonholing, see line $9$ of Algorithm \ref{alg:zombi}.

\textbf{LCB Adaptive} builds off of previous work that also tune sampling based on the number of experiments collected, $n$ \cite{Srinivas2010, phoenics, gryffin}. In this paper, we design LCB Adaptive to tune its hyperparameter to become less explorative as more samples are collected. For example, as the $n$ increases, LCB Adaptive decays its $\beta$ hyperparameter value to become less explorative and more exploitative. Specifically, this information feedback received by the function determines the contribution of both $\mu(X)$ and $\sigma(X)$ to the acquisition value, $a$. Similar to EI Abrupt, LCB Adaptive computes an acquisition value, $a\in[0,1]$, for a given $X$, wherein the $X$ with the highest $a$ is selected by the acquisition function as the next suggested experiment to measure. LCB Adaptive is implemented for a minimization problem as:

\begin{equation}
    a_\textrm{LCB Adaptive}(X, n; \beta, \epsilon) = \mu(X)-\epsilon^n\beta{\sigma(X)},
\end{equation}
where $n$ is the number of experiments sampled, and $\beta=3$ and $\epsilon=0.9$ are hand-tuned initialization hyperparameters selected based on \textit{a priori} domain knowledge of the function's performance on a variety of different problems. Having a large $\beta$ and an $\epsilon$ close to $1$ supports a gradual decay from very explorative to very exploitative, rather than a rapid decay. In the following section (Section \ref{sec:pigeon}), the dynamic EI Abrupt and LCB Adaptive are shown to both discover optima faster and avoid pigeonholing into local minima better than their static counterparts by actively balancing the ratio of exploitation to exploration using learned information about the quality and quantity of previously sampled experiments.

\textbf{EI Abrupt} is a novel implementation that flips between the exploitative EI \cite{ei} and explorative LCB \cite{Auer2002} acquisition functions based on the computed finite differences of recently evaluated experiments. For example, if the evaluated experiment $y$-values plateaus for three or more experiments in a row, EI Abrupt will abruptly switch from a greedy sampling policy to a more explorative sampling policy. Specifically, this information feedback received by the function determines if the current round of sampling should exploit the surrogate mean values, $\mu(X)$, or explore the surrogate variances, $\sigma(X)$. EI Abrupt computes an acquisition value, $a\in[0,1]$, for a given $X$, wherein the $X$ with the highest $a$ is selected by the acquisition function as the next suggested experiment to measure. EI Abrupt is implemented for a minimization problems as:

\begin{align}
\begin{split}
a_\textrm{EI Abrupt}(X, y; \beta, \xi, \eta)=&
\begin{cases}
    \left(\mu(X) - y^* - \xi\right)\Phi(Z)+\sigma(X)\psi(Z), & \text{if } |\Delta \{y_{n-3...n}\}|\leq \eta\\
    \mu(X)-\beta{\sigma(X)}, & \text{otherwise}
\end{cases}\\
Z =& \frac{\mu(X) - y^* - \xi}{\sigma(X)},
\end{split}
\end{align}
where $y^*$ is the lowest measured target value thus far (\textit{i.e.}, the running minimum), $\Phi(\cdot)$ is the cumulative density function of the normal distribution, $\psi(\cdot)$ is the probability density function of the normal distribution, and $|\Delta \{y_{n-3...n}\}|$ is the absolute value of the finite differences of the set of target values of the last three sampled experiments. Moreover, $\beta=0.1$, $\xi=0.1$, and $\eta=0$ are hand-tuned initialization hyperparameters used for the rest of the paper for EI Abrupt. Moreover, for standard LCB and EI, $\beta=1$ and $\xi=0.1$ hyperparameters are used, respectively. These hyperparameters were selected based on \textit{a priori} domain knowledge of EI Abrupt performance on a variety of different problems. The most important hyperparameter for efficient sampling is $\beta$, whose ideal value is non-obvious, but it is found that  $\beta=0.1$ allows EI Abrupt to switch into an explorative sampling policy while still having a strong weight on the surrogate means, implying that exploration does not veer far.

\section{Demonstration of \texttt{ZoMBI} Mechanics\deleted{ using an Ackley Function}}
\subsection{Zooming Bounds}

\begin{figure}[h!]
\centering
\begin{subfigure}[b]{0.9\textwidth}  
\includegraphics[width=\textwidth]{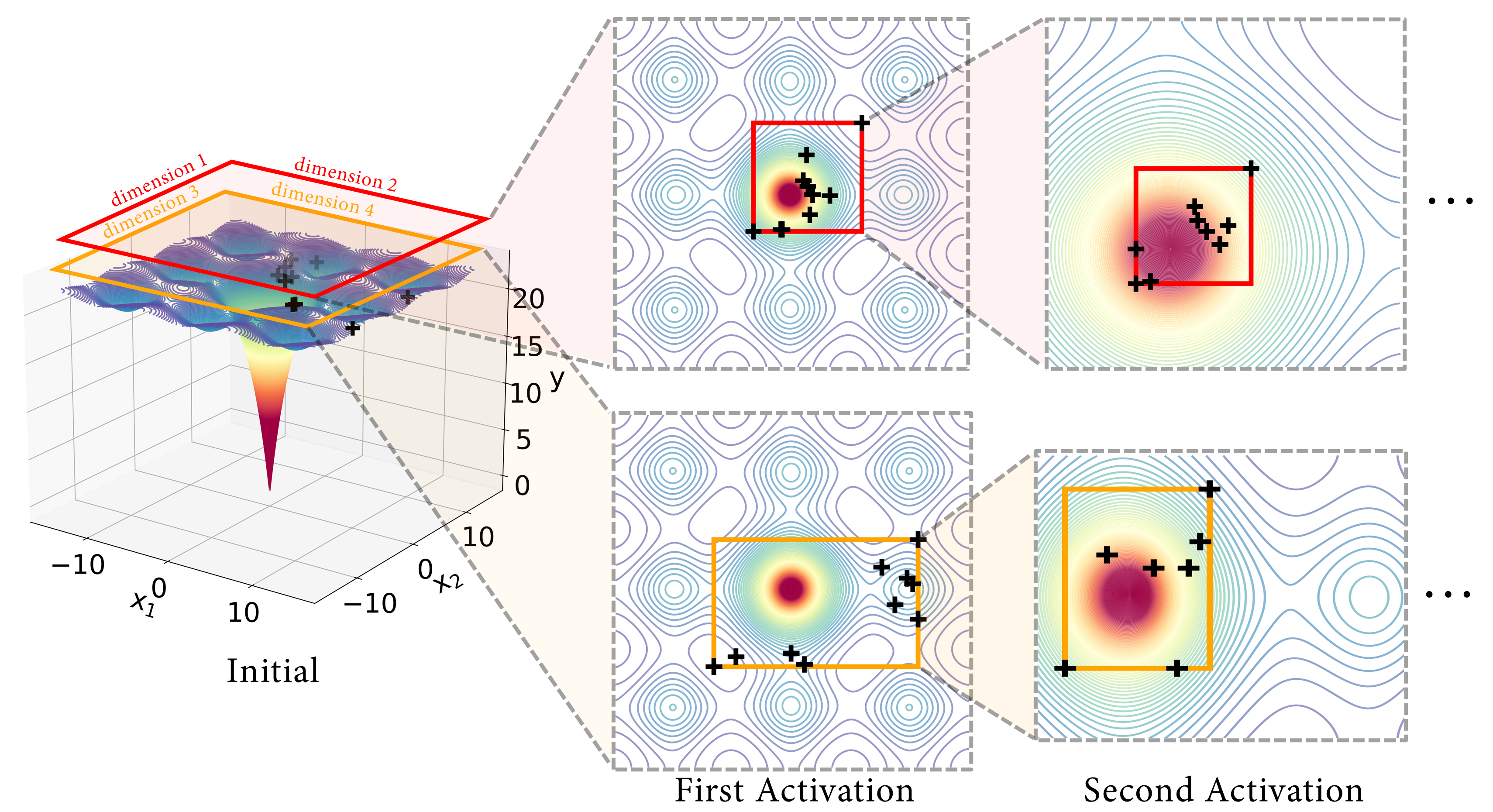}
\end{subfigure}\hfill%
\caption{\textbf{Zooming Search Bounds.} For every activation of \texttt{ZoMBI}, the search bounds are zoomed inward based on the prior best-performing memory points. A 4D Ackley function manifold is projected in 2D. The bounding regions of each 2D slice are illustrated by the red and orange boxes. The $\phi$ number forward experiments sampled for each activation, $\alpha$, are illustrated as black markers. The global optimum is indicated by the red region of the heatmap.}
\label{fig:zombi}
\end{figure}

Zooming in the search bounds on the manifold addresses challenge number one of optimizing NiaH problems, which is the challenge of finding the general hypervolume region that contains the needle-like optimum. Figure \ref{fig:zombi} illustrates how the \texttt{ZoMBI} algorithm iteratively zooms in the search bounds based on the number of activations, $\alpha$. An Ackley function is used as a simulated example due to its non-convexity and needle-like global optimum \cite{ackley1987, Adorio2005}. For each activation, $m$ prior points that achieved the lowest target values, $y$, are retained in memory and used to zoom the search bounds in. This zooming occurs independently across each dimension and is based on the minimum and maximum values of the $m$ memory points along each dimension, as shown in Equation \ref{eq:bounds}. The red and orange rectangles illustrate the evolution of the bounds over space and time. Initially, sampling occurs across the entire manifold for $\phi$ forward experiments per activation, shown by the black markers. However, by using the best-performing memory points to zoom in the search bounds, pigeonholing into local minima can also be avoided as the search bounds are pulled away from these trap minima and move closer towards the global minimum basin of attraction. The iterative zooming of \texttt{ZoMBI} does not guarantee convergence on the global optimum, but if a sufficient initialization set is obtained, convergence often gets close to the global optimum as shown across several examples in Figure \ref{fig:search} and Figures \ref{fig:poisson_bench}, \ref{fig:zt_bench}, \ref{fig:fire_bench}. Furthermore, we comprehensively demonstrate the performance limitations of \texttt{ZoMBI} where initializations miss extreme needle-like optima in Figure \ref{fig:hypervol} and where optima are near the edges of a manifold in Figure \ref{fig:distance}.

\subsection{Memory Pruning}

\label{sec:compute}
\begin{figure}[h!]
\centering
\begin{subfigure}[b]{0.65\textwidth}  
\includegraphics[width=\textwidth]{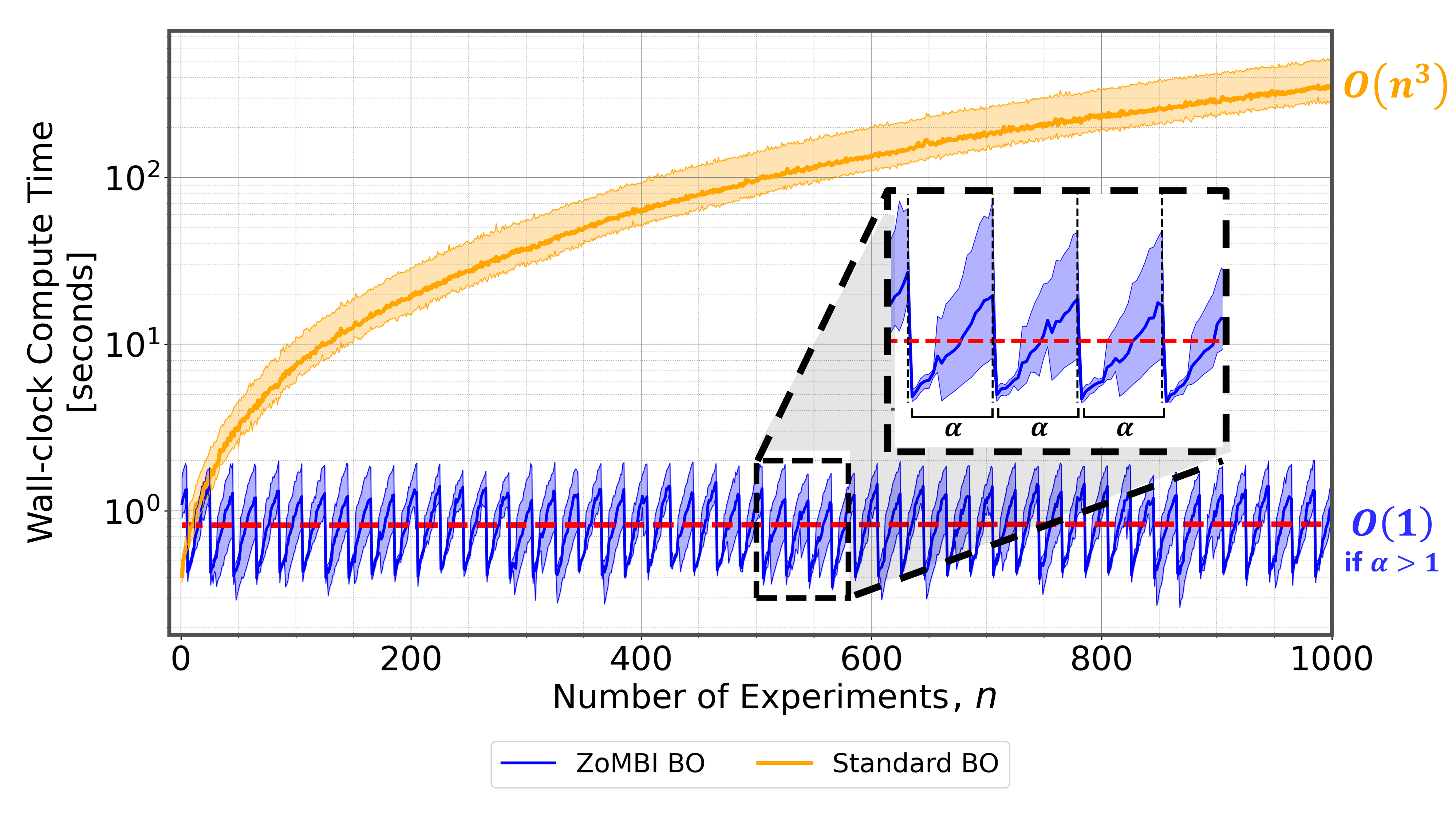}
\end{subfigure}\hfill%
\caption{\textbf{Wall-clock Compute Time.} The compute time per experiment is illustrated for traditional BO with a GP surrogate (orange) and for \texttt{ZoMBI} with a GP surrogate (blue) with the $y$-axis in log-scale. Four independent trials of each method were run to optimize a 6D Ackley function with a narrow basin of attraction using an NVIDIA Tesla Volta V100 GPU \cite{reuther2018interactive}. Each trial of standard BO and \texttt{ZoMBI} is run using one of the four acquisition functions: LCB, LCB Adaptive, EI, and EI Abrupt. The averages of the trials are shown as solid orange and blue lines while the shaded regions indicate the maximum and minimum compute times bounds. The red dashed line indicates the trend of the \texttt{ZoMBI} compute times. The measured compute time includes the time to compute the GP surrogate model and the time to acquire an experiment from the surrogate.}
\label{fig:compute}
\end{figure}

As more experiments are amassed and committed to memory to run traditional BO by computing the GP surrogate, the compute time increases polynomially, following the $O(n^3)$ time complexity of GP matrix inversion \cite{BelyaevMikhail2014, snelson2005, Rasmussen2005, Correa2006, Bui2017, Lan2020}. This complexity is unfavorable as it leads to compounding compute times as more experiments are run. Therefore, we implement a memory pruning feature into the \texttt{ZoMBI} algorithm that iteratively selects which prior data points to keep and which to prune from the memory during each activation, $\alpha$. Memory pruning is demonstrated to remove redundant features during the optimization procedure. Figure \ref{fig:prune} illustrates how \texttt{ZoMBI} accelerates the convergence of a GP prediction to the precise location of the true. However, only data within the newly computed bounds of \texttt{ZoMBI} are used prediction of the true target, hence, all data outside this boundary becomes redundant and is pruned to decrease compute time. 

Through memory pruning, the number of experiments used to train the GP surrogate varies between $[i,i+\phi]$ for every $\alpha$, rather than being proportional to $n$, where the number of initialization samples is fixed at $i=5$. In this paper, we use $\phi\in[0,10]$, \textit{i.e.}, once $\phi=10$, the activation is complete and resets to $\phi=0$. This is computationally favorable because $\{X_i\} \cup \{X_\phi\} \subseteq \{X_n\}$. Thus, for a single $\alpha$, the time complexity is $O((i+\phi)^3) \approx O(\phi^3)$, since $i$ is fixed. Furthermore, since the range of $\phi$ is capped, a non-increasing sawtooth pattern in compute time is exhibited, illustrated in Figure \ref{fig:compute}. Therefore, the compute complexity of \texttt{ZoMBI} trends towards $O(1)$ for $\alpha>1$ as a result of the efficient memory pruning process. After collecting $1000$ experiments, the compute time of traditional BO trends towards $>400$ seconds per experiment, whereas for \texttt{ZoMBI} the compute time maintains a constant trend of approximately 1 second per experiment. Therefore, the memory pruning feature of \texttt{ZoMBI} accelerates the optimization compute time by over $400$x at $n=1000$ and achieves further relative acceleration as $n$ increases.

\subsection{Anti-pigeonholing}
\label{sec:pigeon}

\begin{figure}[h!]
\centering
\begin{subfigure}[b]{0.24\textwidth} 
\includegraphics[width=\textwidth]{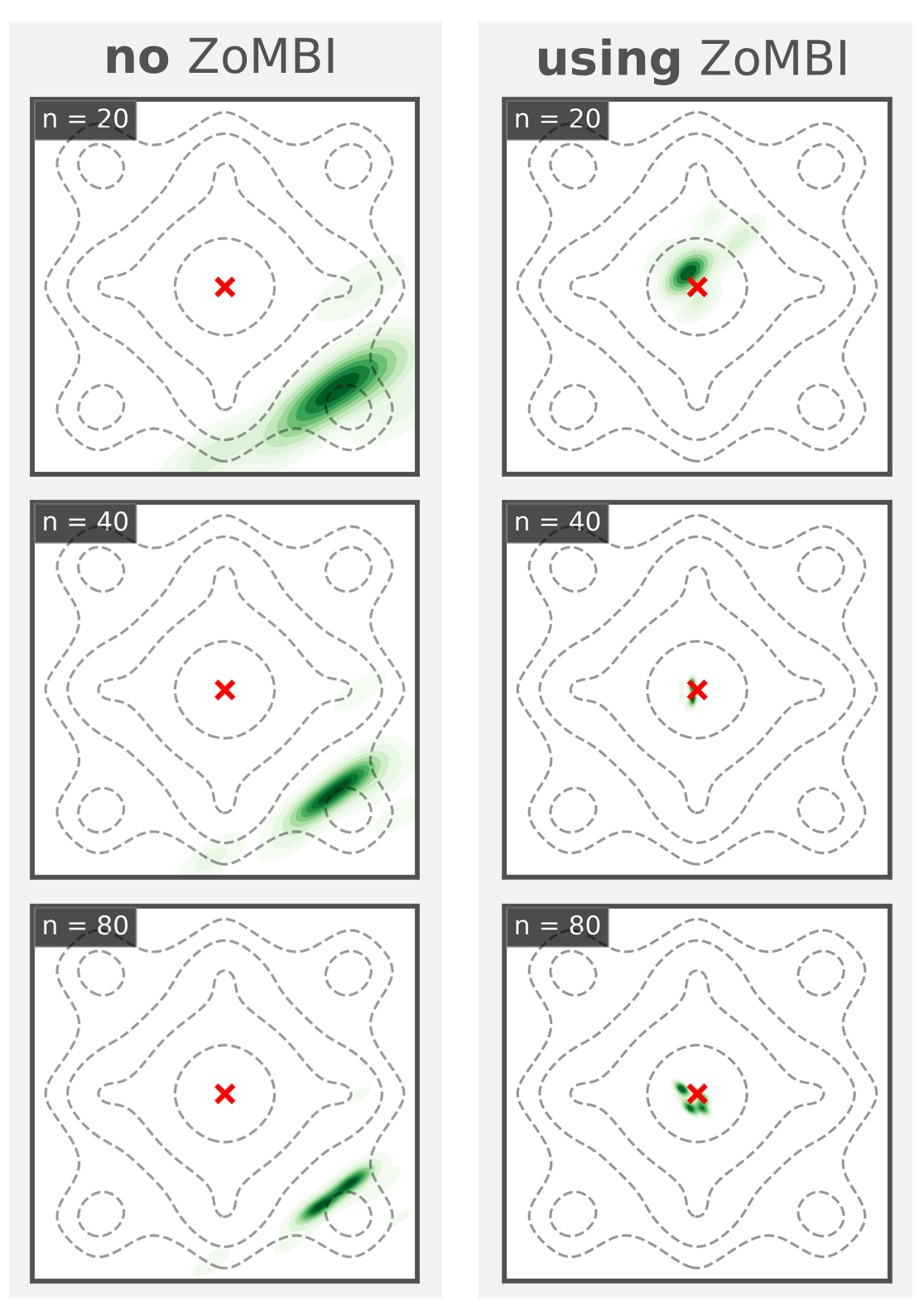}
\caption{LCB}
\end{subfigure}\hfill%
\begin{subfigure}[b]{0.24\textwidth} 
\includegraphics[width=\textwidth]{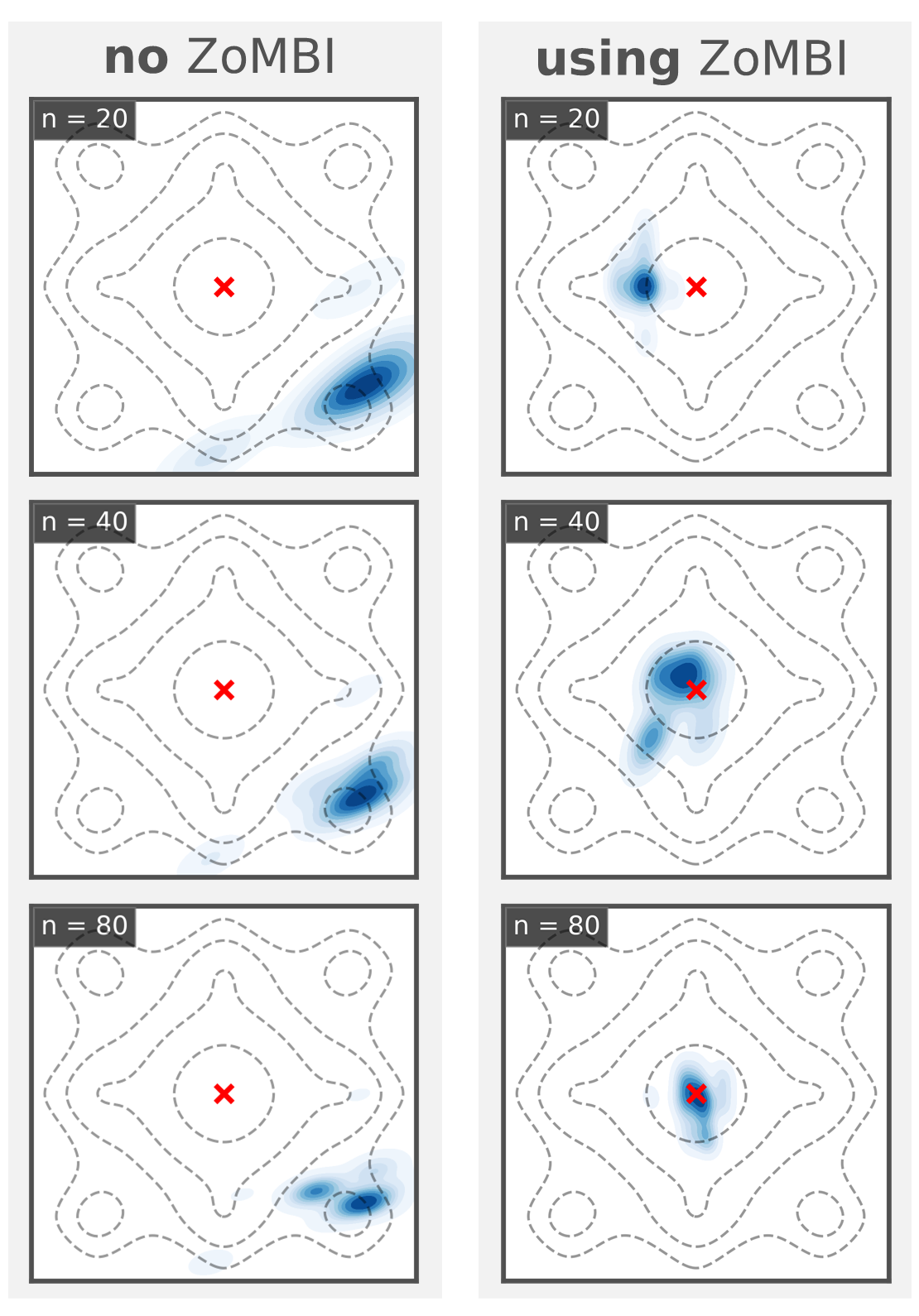}
\caption{LCB Adaptive}
\end{subfigure}\hfill%
\begin{subfigure}[b]{0.24\textwidth}  
\includegraphics[width=\textwidth]{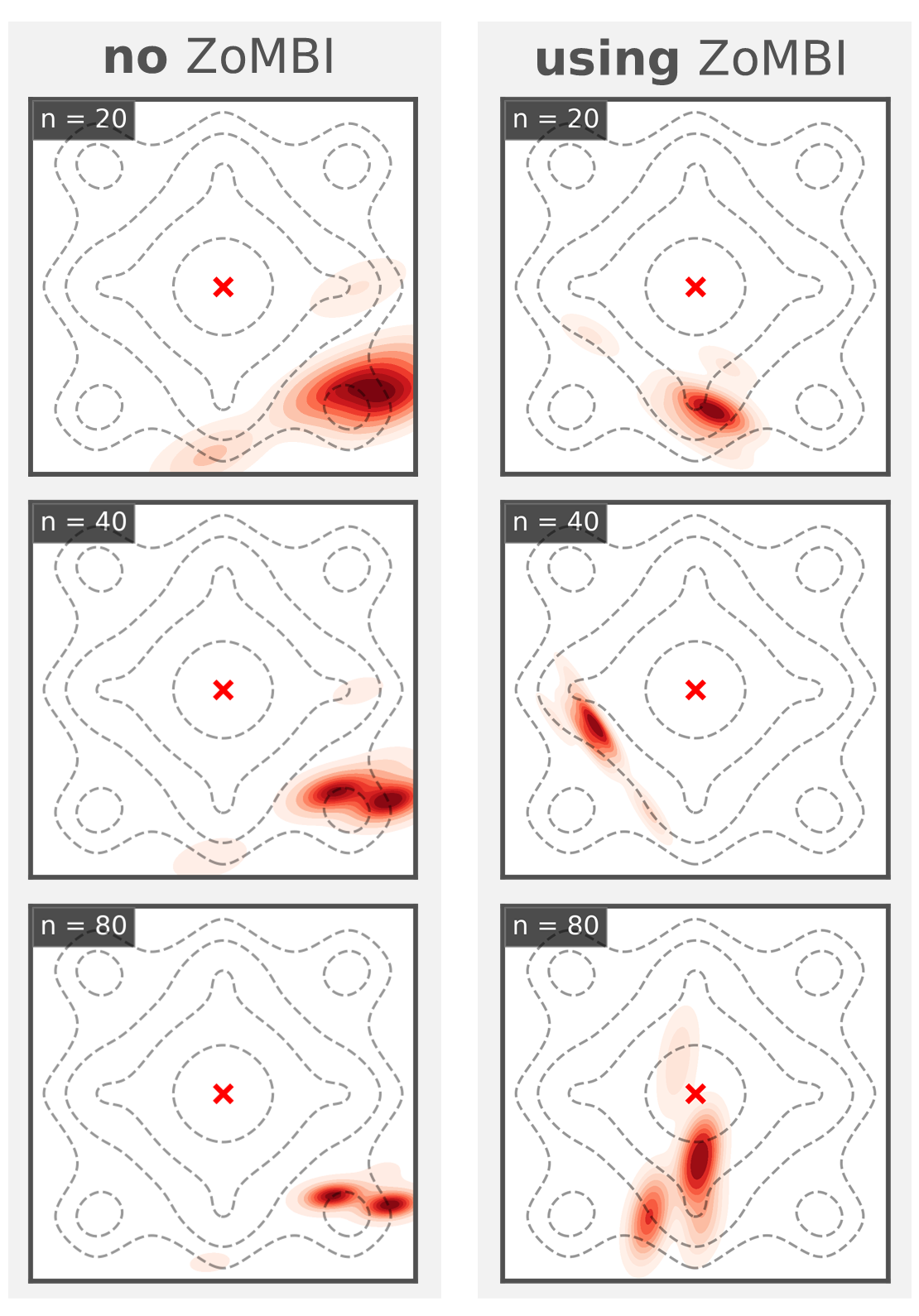}
\caption{EI}
\end{subfigure}\hfill%
\begin{subfigure}[b]{0.24\textwidth}  
\includegraphics[width=\textwidth]{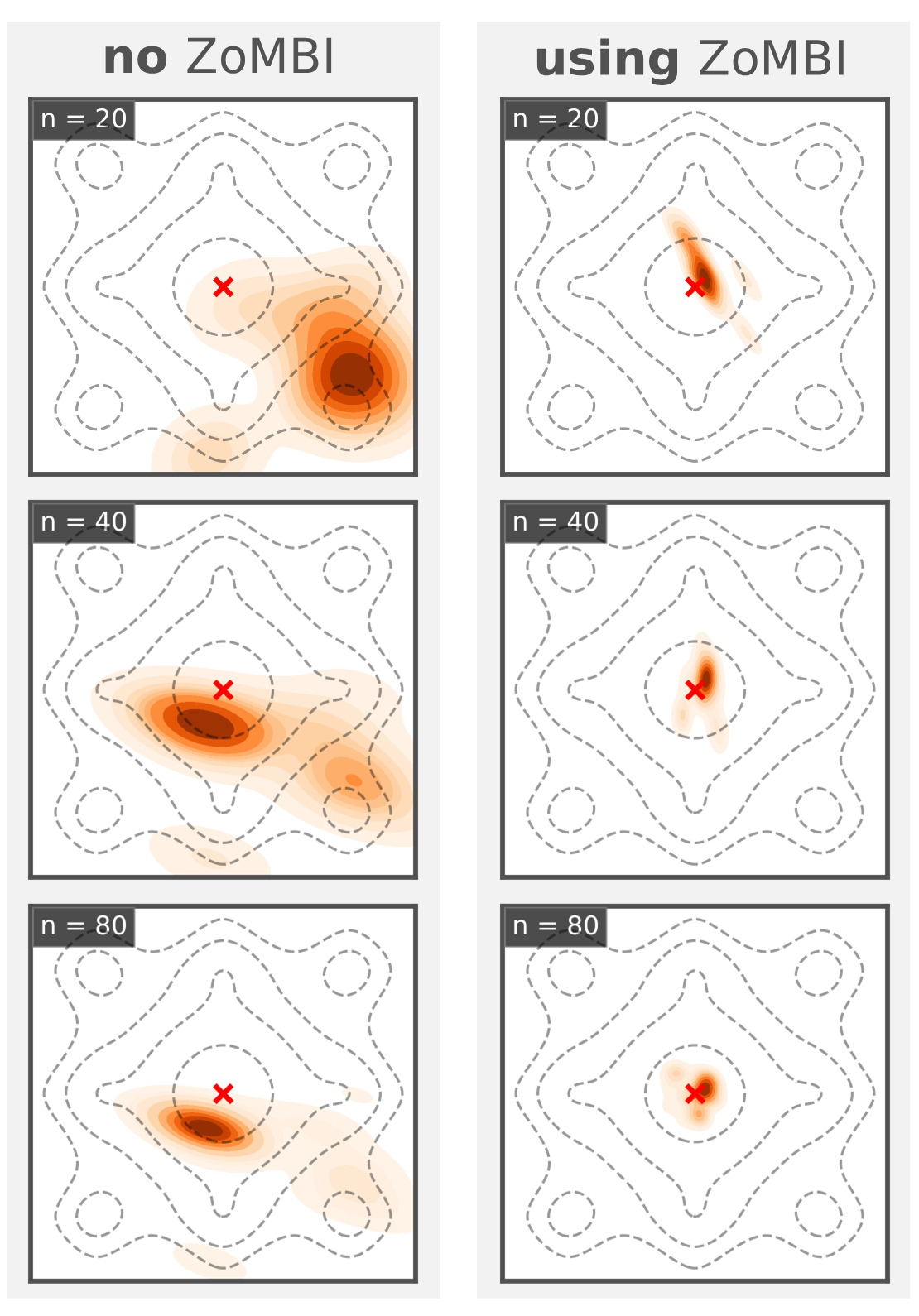}
\caption{EI Abrupt}
\end{subfigure}\hfill%
\caption{\textbf{Acquisition Function Sampling Density.} The colored heatmaps indicate the regions of a 2D slice from a 6D Ackley function where sampling density is high for each respective acquisition function: (a) LCB, (b) LCB Adaptive, (c) EI, and (d) EI Abrupt. The contour lines indicate the manifold topology with local minima as the circular and pointed regions of the contours. The red "x" indicates the global minimum. For each acquisition function, the left panel shows the sampling density after $n=\{20,40,80\}$ evaluated experiments without the use of \texttt{ZoMBI} while the right panel shows the sampling density after $n=\{20,40,80\}$ evaluated experiments with the use of \texttt{ZoMBI}.}
\label{fig:search}
\end{figure}

Pigeonholing into the local minima of a function occurs when an optimization algorithm has insufficient learned knowledge of the manifold topology to continue exploring potentially profitable regions or when the algorithm's hyperparameters are improperly tuned, leading to overly exploitative tendencies \cite{Liu2022, Liang2021}. The \texttt{ZoMBI} algorithm's anti-pigeonholing capabilities are two-fold: (1) the zooming search bounds help the acquisition function to quickly stop sampling local minima once a better performing data point is found and (2) actively learned acquisition function hyperparameters use knowledge about the domain to help exit a local minimum. Figure \ref{fig:search} demonstrates the anti-pigeonholing capabilities of \texttt{ZoMBI} on optimizing a 6D Ackley function with both static and dynamic acquisition functions, compared to that of traditional BO. 

The needle-like global minimum is indicated by the red "x" and the local minima are indicated by the circular and pointed regions of the contour lines. The sampling density of each acquisition function is illustrated by the heatmap, where the darker colors indicate higher sampling density regions. The goal is to get high sampling density near the red "x". It is shown that without \texttt{ZoMBI} being activated, the LCB, LCB Adaptive, and EI acquisition functions all end up pigeonholing into local minima. However, EI Abrupt initially pigeonholes into a local minima but then switches from an exploitative to an explorative mode to jump out of the local minimum and converge closer to the global. Conversely, when running the optimization procedure with \texttt{ZoMBI} active, all of the acquisition functions except the most exploitative, EI, converge onto the global minimum fast. LCB Adaptive and EI are shown to initially start sampling towards a local minima, but as \texttt{ZoMBI} is iteratively activated, the search bounds zoom in closer to the global minimum. Thus, with the combination of dynamic acquisition functions and zooming search bounds, pigeonholing into sub-optimal local minima can be more readily avoided while optimizing NiaH problems, although avoidance is not guaranteed, as shown by the sampling density of EI. The combination of the three foundational features of \texttt{ZoMBI}, (1) zooming bounds, (2) memory pruning, and (3) anti-pigeonholing drives fast optimization of NiaH problems and in most cases, does not sacrifice the ability to converge on the global optimum.

\section{Experiments}

\subsection{Large-scale Optimum Hypervolume Analysis via Ackley Permutations}

Before assessing the performance of \texttt{ZoMBI} on the three real-world datasets, we use 144 permutations of the Ackley function to stress-test the capability of \texttt{ZoMBI} to discover the global optimum basin of attraction, given two varying dataset hyperparameters: (1) basin of attraction width and (2) dimensionality. The basin of attraction hypervolume is determined by both the width of the basin and the dimensionality of the manifold, hence, as the basin becomes narrower in width and as the dimensionality increases, the percentage of hypervolume space taken up by the basin decreases, \textit{i.e.} the optimum becomes more needle-like. The Ackley permutations have varying basin hypervolumes from 0.001\% to 100\% and varying manifold dimensionalities from 2D to 10D. For this experiment, we aim to determine types of manifold topologies that \texttt{ZoMBI} best optimizes while quantifying those limits with the Pareto front.

\begin{figure}[h!]
\centering
\begin{subfigure}[b]{0.9\textwidth}  
\includegraphics[width=\textwidth]{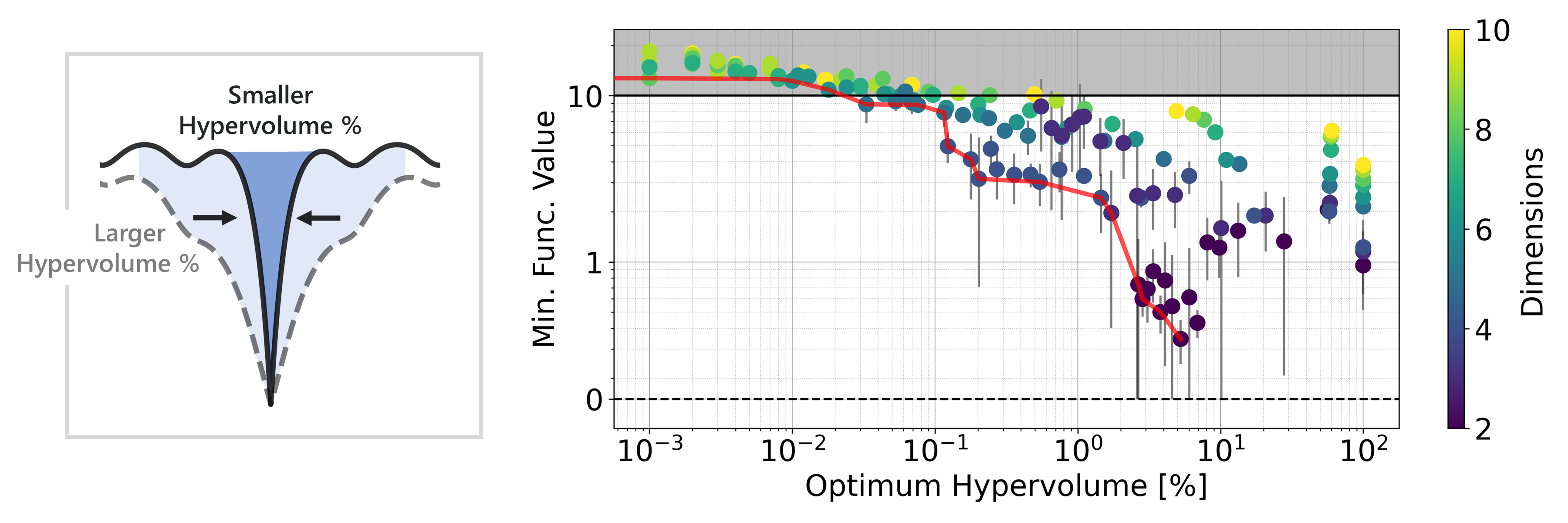}
\end{subfigure}\hfill%
\caption{\textbf{Varying Optimum Hypervolume.} (left) Depiction of decreasing optimum basin of attraction hypervolume in 1D. (right) The Pareto-optimal dataset hyperparameters for usage with the \texttt{ZoMBI} algorithm over 144 analytical datasets with 48 independent trials each: 12 trials for each of the four acquisition functions, LCB, LCB Adaptive, EI, and EI Abrupt, for a total of $6912$ independent trials. Each analytical dataset is a permutation of the Ackley function with a different optimum basin of attraction width and manifold dimensionality. Hypervolume percent makeup is synthetically decreased both by decreasing the basin of attraction width and by increasing the manifold dimensionality. Each scatter point represents the median final minimum function evaluation after $1000$ experiments across the 48 independent trials initialized with a fixed set of $i=5$ samples. The colorbar of the scatter point represents the dimensionality of the manifold tested and the error bars represent the variance across the 48 trials. The possible function values for every dataset vary between $[0,25]$, hence, for the Ackley function as further described in the supplemental Section \ref{sec:ackley}, trials achieving minimum function values $<10$ are considered to have found the optimum basin of attraction while trials with function values $\geq 10$ after $1000$ experiments are considered to be trapped in local minima. Both the $x$- and $y$-axes are in log-scale.}
\label{fig:hypervol}
\end{figure}

Figure \ref{fig:hypervol} shows the results of this large-scale optimization experiment of 48 independent trials of \texttt{ZoMBI} across each of the 144 unique permutations of the Ackley function dataset with varying optimum hypervolumes and dimensionality. All points below the grey-shaded region fall within the optimum basin of attraction. The red trace of the Pareto front indicates the narrowest optimum hypervolume and dimensionality conditions of a dataset that result in the best minimum function value being discovered. We show that with an initialization set of $i=5$, \texttt{ZoMBI} can reliably discover the global minimum region for needles as narrow as $0.05\%$ of total hypervolume space. Moreover, as the optimum becomes narrower than $0.05\%$ of the total hypervolume, the initialization set is no longer sufficient and \texttt{ZoMBI} gets trapped in local minima, as indicated by the greyed-out region. Conversely, as the optimum becomes wider than $5\%$ of the total hypervolume, the manifold becomes flatter, expressing the greedy nature of \texttt{ZoMBI} to falsely zoom inward to less ideal function values than it would for narrower optimum conditions. This experiment quantifies the range of \texttt{ZoMBI}'s Goldilocks zone to be between $0.05\%$ and $5\%$ optimum hypervolume. Therefore, for ideal performance, \texttt{ZoMBI} is best used on datasets with optimum conditions consisting of between $0.05\%$ and $5\%$ of the total number of conditions. This optimum hypervolume trade-off of \texttt{ZoMBI} is further assessed relative to other optimization methods in Figure \ref{fig:smooth}. 

\begin{figure}[h!]
\centering
\begin{subfigure}[b]{1\textwidth}  
\includegraphics[width=\textwidth]{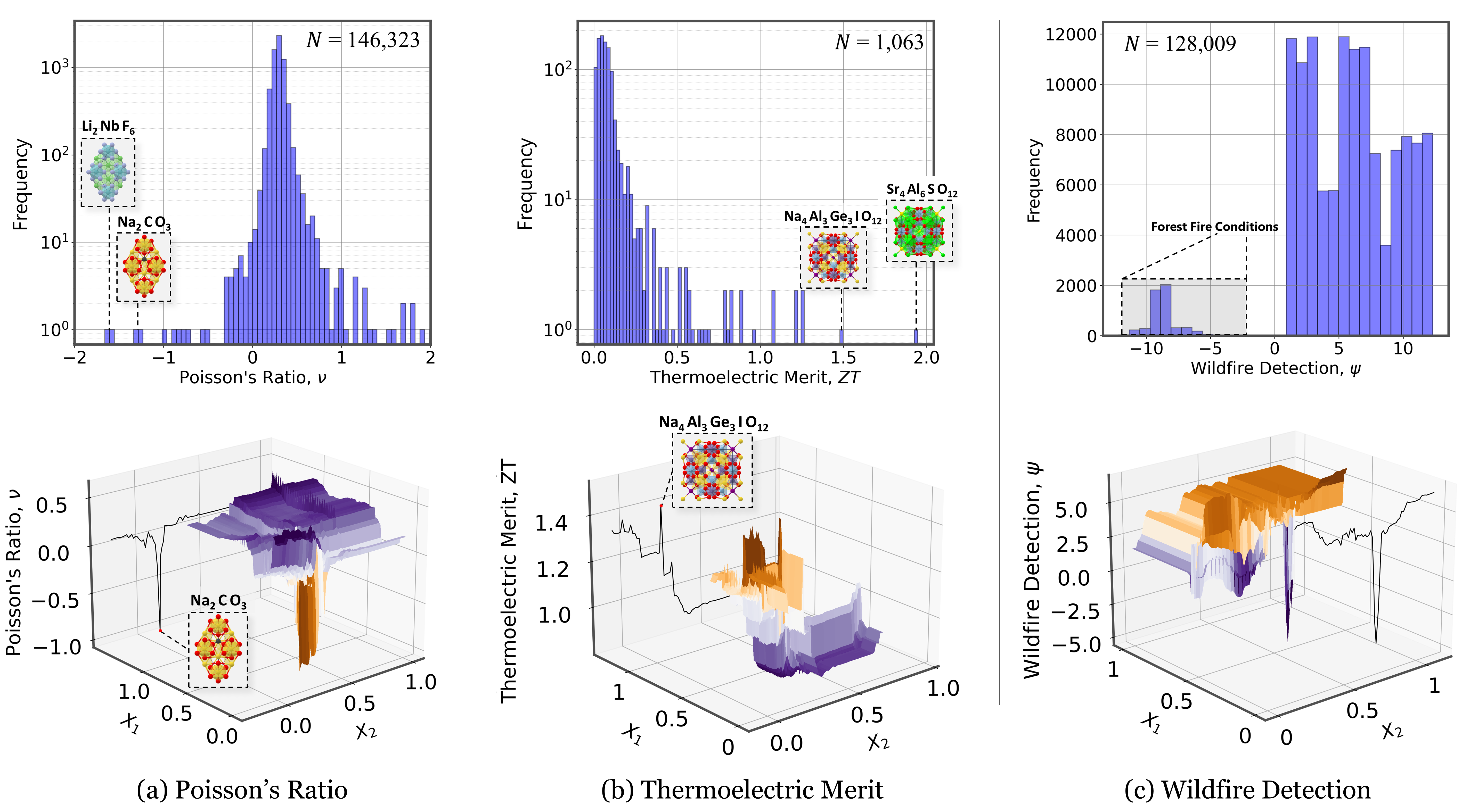}
\end{subfigure}\hfill%
\caption{\textbf{Data Distributions of Real-world Needle-in-a-Haystack Datasets} (top) The histogram distributions of the full real-world datasets with callouts for optimum conditions: (a) Poisson's Ratio with $146$k materials in the dataset and $\nu_\textrm{min}=\{-1.7,-1.2\}$, (b) Thermoelectric Figure of Merit with $1$k materials in the dataset computed by BoltzTraP \cite{Madsen2006} and $ZT_\textrm{max}=\{1.4,1.9\}$, (c) Wildfire Detection with $128$k meteorological conditions collected over 33 months from January 2018 to September 2020 from CIMIS \cite{Kohli2020} and $\psi < 0$ conditions indicating those with a high likelihood of wildfire outbreaks. (bottom) The noisy, non-convex manifold topologies of each dataset generated by a random forest regression with $500$ trees. Each manifold is a projected 3D slice of higher dimensional space with the $z$-axis and colorbar indicating the target property, where (a) $X_1$ is density and $X_2$ is formation energy, (b) $X_1$ is formation energy and $X_2$ is band gap, (c) $X_1$ is evapotranspiration and $X_2$ is precipitation.}
\label{fig:all_hists}
\end{figure}


In the next section, three real-world datasets are optimized using \texttt{ZoMBI} -- each of these datasets have extreme data imbalances, illustrated in \ref{fig:all_hists} within the specified ideal ranges of \texttt{ZoMBI} performance. The 6D Poisson's Ratio dataset has an imbalance of $0.82\%$ optimum conditions, the 6D Thermoelectric Figure of Merit dataset has an imbalance of roughly $1.32\%$ optimum conditions, and the 11D wildfire detection dataset has an imbalance of $4.16\%$ optimum conditions. This range of ideal performance of \texttt{ZoMBI} between $0.05\%$ and $5\%$ optimum hypervolume is facilitated by the initialization set. Hence, to improve performance for narrower optima, either the number of initialization samples must be increased, or initialization conditions should be adjusted. Additional initialization conditions experiments of \texttt{ZoMBI} are shown in supplemental Section \ref{sec:edge}.

\subsection{6D Poisson's Ratio}
\label{sec:poisson}

\begin{figure}[h!]
\centering
\begin{subfigure}[b]{0.9\textwidth} 
\includegraphics[width=\textwidth]{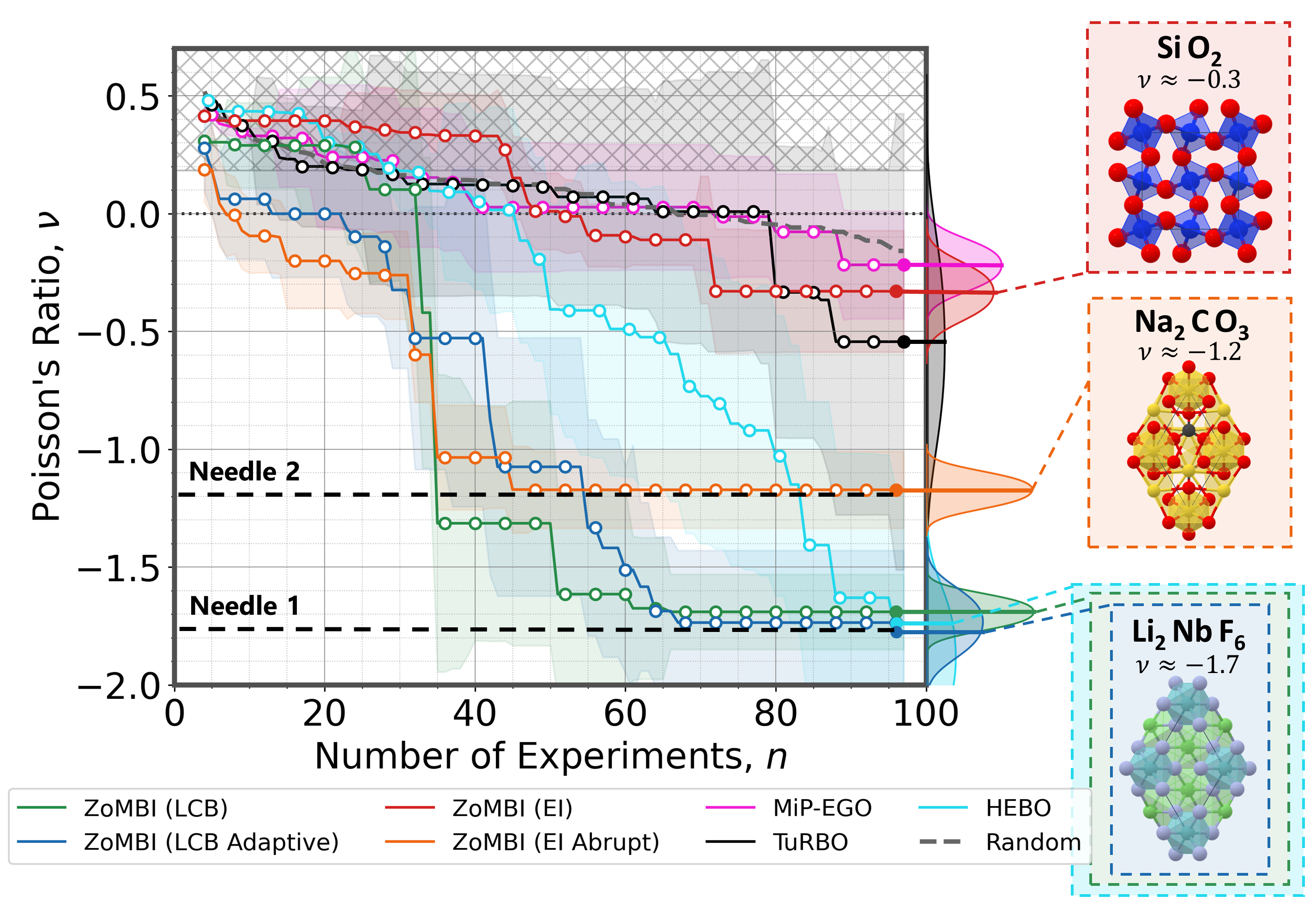}
\end{subfigure}\hfill%
\caption{\textbf{Discovery of Rare Negative Poisson's Ratio Materials.} The optimization objective is to find the material with the minimum Poisson's ratio in $100$ experiments from the dataset presented in Figure \ref{fig:all_hists}(a). The green, blue, red, and orange lines indicate the median best running evaluated sample of \texttt{ZoMBI} using the LCB, LCB Adaptive, EI, and EI Abrupt acquisition functions, respectively. The pink, black, and teal lines indicate the median best running evaluated sample of the methods \texttt{MiP-EGO}, \texttt{TuRBO}, and \texttt{HEBO} respectively. Random sampling is illustrated as a dashed grey line for benchmarking. The median for each method is taken over the best $12$ independent model runs. The shaded regions indicate the variance between model runs. The cross-hatched region indicates the space discovered by standard BO methods, without the use of \texttt{ZoMBI}, which use the same hyperparameters as described in Section \ref{sec:acquisition}. The distribution across all $12$ model runs of the final sampled experiment for each method is shown as a kernel density estimation (KDE) along the $y$-axis. The y-values for the needle-like optima are indicated by dashed black lines.}
\label{fig:poisson_bench}
\end{figure}

The first experimental dataset is 6-dimensional and consists of 146k materials from the publicly available Materials Project database with different mechanical properties, described by Poisson's Ratio, $\nu$ \cite{Jain2013}. Only $0.82\%$ of the total 146k materials have a negative Poisson's Ratio, $\nu < 0$ \cite{Dagdelen2017, Jain2013, DeJong2015, Saxena2016}. Hence, for this experiment, we aim to \textbf{minimize} $\mathbf{\nu}$. A positive $\nu > 0$, describes a material that expands when a compressive load is applied to the orthogonal direction \cite{Belyadi2019, Poplavko2019}. Conversely, a negative $\nu < 0$ describes a material that contracts rather than expands when compressed in the orthogonal direction, denoted as an auxetic material -- a rare phenomenon \cite{Dagdelen2017, Lakes2008}. Auxetic materials with highly negative Poisson's ratios have energy absorptive properties that are ideal materials for wearable medical devices and protective armor that must absorb the energy of large impacts to keep bones from shifting or to inhibit the penetration of the protective layer \cite{Saxena2016, Liu2006}.

Figure \ref{fig:poisson_bench} demonstrates the optimization performance of \texttt{ZoMBI} on the Poisson's Ratio dataset compared to \texttt{MiP-EGO}, \texttt{TuRBO}, and \texttt{HEBO}. The \texttt{ZoMBI} algorithm is run with each of the four acquisition functions: LCB, LCB Adaptive, EI, and EI Abrupt. In under $100$ evaluated experiments, LCB and LCB Adaptive discover the global minimum NiaH material, Li$_2$NbF$_6$ ($\nu \approx -1.7$). The variance of $\nu$ values for the final experiment across all ensemble runs is illustrated as a KDE plot for each method to highlight the sampling density and general rate of success. \texttt{HEBO} discovers the global minimum after \texttt{ZoMBI} with LCB and LCB Adaptive, however, the spread of runs for \texttt{ZoMBI} is narrower than that of \texttt{HEBO}, which indicates that for this problem, \texttt{ZoMBI} can more consistently discover the minimum, that is 3x lower than those discovered by \texttt{MiP-EGO} and \texttt{TuRBO}. Furthermore, the rate of convergence on Needle 1 is faster for \texttt{ZoMBI} than \texttt{HEBO}.

Figure \ref{fig:all_hists}(a) illustrates the distribution of $\nu$ values within the full dataset. The ground truth "needle" materials with the lowest $\nu$ values are Li$_2$NbF$_6$ with $\nu \approx -1.7$ and Na$_2$CO$_3$ with $\nu \approx -1.2$. \texttt{ZoMBI} with the LCB and LCB Adaptive acquisition functions and \texttt{HEBO} discover Li$_2$NbF$_6$, while \texttt{ZoMBI} with the EI Abrupt acquisition function discovers Na$_2$CO$_3$.

\subsection{6D Thermoelectric Figure of Merit}

\begin{figure}[h!]
\centering
\begin{subfigure}[b]{0.9\textwidth} 
\includegraphics[width=\textwidth]{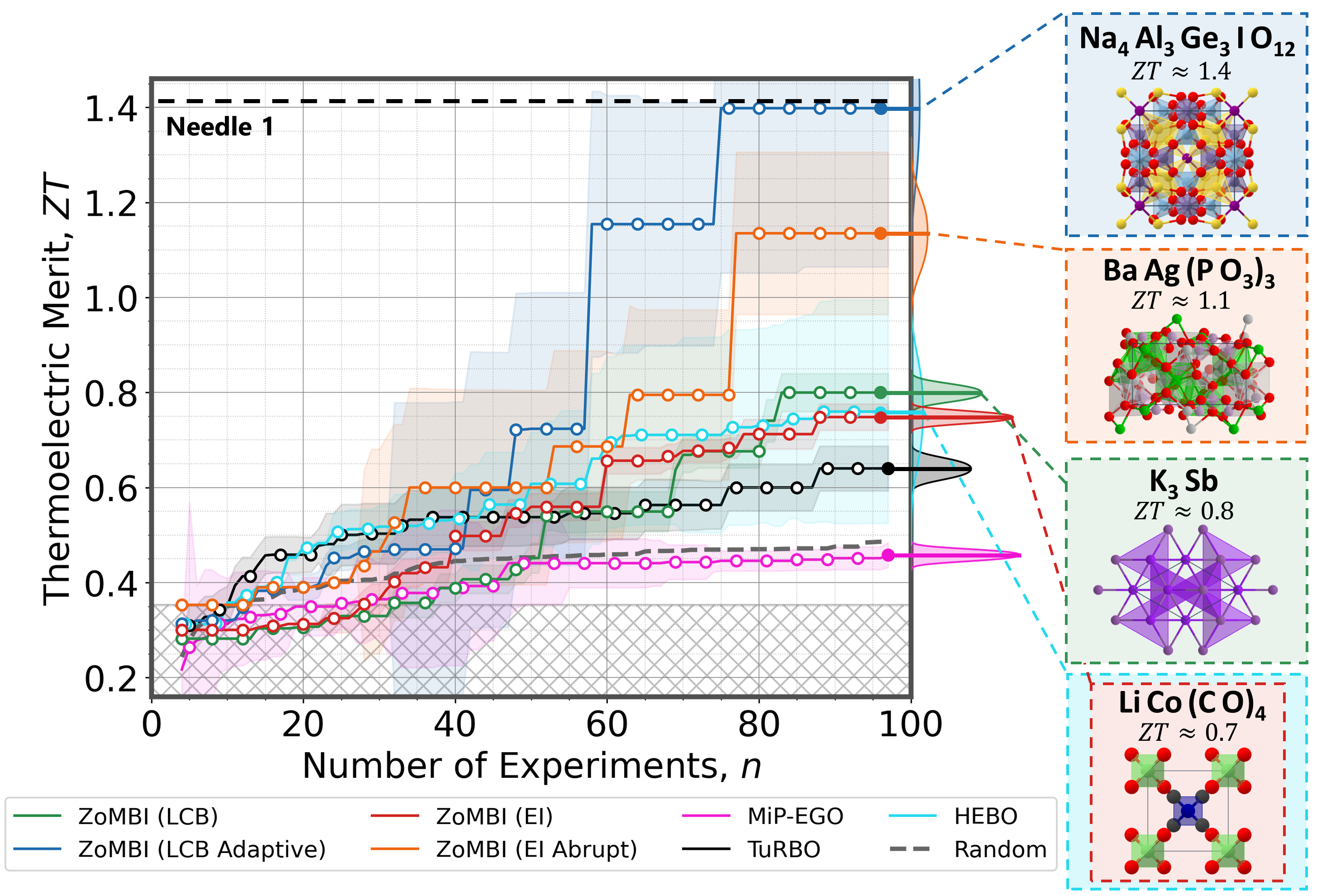}
\end{subfigure}\hfill%
\caption{\textbf{Discovery of Rare Positive Thermoelectric Figure of Merit Materials.} The optimization objective is to find the material with the maximum Thermoelectric Figure of Merit in $100$ experiments from the dataset presented in Figure \ref{fig:all_hists}(b). The green, blue, red, and orange lines indicate the median best running evaluated sample of \texttt{ZoMBI} using the LCB, LCB Adaptive, EI, and EI Abrupt acquisition functions, respectively. The pink, black, and teal lines indicate the median best running evaluated sample of the methods \texttt{MiP-EGO}, \texttt{TuRBO}, and \texttt{HEBO} respectively. Random sampling is illustrated as a dashed grey line for benchmarking. The median for each method is taken over the best $12$ independent model runs. The shaded regions indicate the variance between model runs. The cross-hatched region indicates the space discovered by standard BO methods, without the use of \texttt{ZoMBI}, which use the same hyperparameters as described in Section \ref{sec:acquisition}. The distribution across all $12$ model runs of the final sampled experiment for each method is shown as a kernel density estimation (KDE) along the $y$-axis. The y-values for the needle-like optima are indicated by dashed black lines.}
\label{fig:zt_bench}
\end{figure}

The second experimental dataset is 6-dimensional and consists of 1k materials with different thermal and electrical properties, described by the Thermoelectric Figure of Merit, $ZT$. Since $ZT$ values are always positive, there is no clear cutoff for what "optimum" conditions are, but with a threshold of $ZT>0.8$, $1.32\%$ of the total 1k materials are considered optimum. A higher $ZT$ indicates that the material is better able to convert a thermal gradient into an electrical current \cite{Hinterleitner2019}. Hence for this experiment, we aim to \textbf{maximize $\mathbf{ZT}$}. Unlike Poisson's Ratio, Thermoelectric Merit is determined by a combination of several variables, rather than a single variable \cite{Hinterleitner2019}: 

\begin{equation}
    ZT=\frac{S^2\sigma}{\kappa}T,
\label{eq:zt}
\end{equation}

where $S$ is the Seebeck coefficient, $\sigma$ is electrical conductivity, $T$ is the average temperature, and $\kappa$ is thermal conductivity. The $ZT$ is computed for each material with valid thermal and electrical properties in the Materials Project database using BoltzTraP \cite{Madsen2006}. $ZT$ is a common figure of merit used to describe the thermal-to-electrical or electrical-to-thermal conversion efficiency of thermoelectric materials \cite{Kim2015, Chen2016, Goldsmid2014, Rodrigo2019}. Materials with high $ZT$ values have a range of applications from usage as solid-state cooling devices to being used as sensors that when heated up, will produce an electrical signal \cite{Salah2020, He2018, Mao2020}.

Figure \ref{fig:zt_bench} demonstrates the optimization performance of \texttt{ZoMBI} on the Thermoelectric Figure of Merit dataset compared to \texttt{MiP-EGO}, \texttt{TuRBO}, and \texttt{HEBO}. In this experiment, although none of the tested methods are able to discover the maximum needle, LCB Adaptive discovers the second highest needle-in-a-haystack material, Na$_4$Al$_3$Ge$_3$IO$_{12}$ ($ZT \approx 1.4$) in under $100$ experiments. Neither \texttt{HEBO}, \texttt{TuRBO}, nor \texttt{MiP-EGO} are capable of discovering any needle-like $ZT$ optima and \texttt{MiP-EGO} performs worse than random sampling in this experiment. The wide variance across runs for \texttt{ZoMBI} and \texttt{HEBO}, shown in the KDE plots, indicate that both methods operate relatively explorative to discover maxima in this topology. Ultimately, this experiment demonstrates that \texttt{ZoMBI} can optimize material objective functions that have a complex combination of variables (Equation \ref{eq:zt}) with roughly 2x better performance than \texttt{HEBO}.

Figure \ref{fig:all_hists}(b) illustrates the distribution of $ZT$ values within the full dataset. The ground truth "needle" materials with the highest $ZT$ values are Sr$_4$Al$_6$SO$_{12}$ with $ZT \approx 1.9$ and Na$_4$Al$_3$Ge$_3$IO$_{12}$ with $ZT \approx 1.4$. \texttt{ZoMBI} with the LCB Adaptive acquisition function is the only method that discovers one of these needles, Na$_4$Al$_3$Ge$_3$IO$_{12}$.

\subsection{11D Wildfire Detection}

\begin{figure}[h!]
\centering
\begin{subfigure}[b]{0.75\textwidth} 
\includegraphics[width=\textwidth]{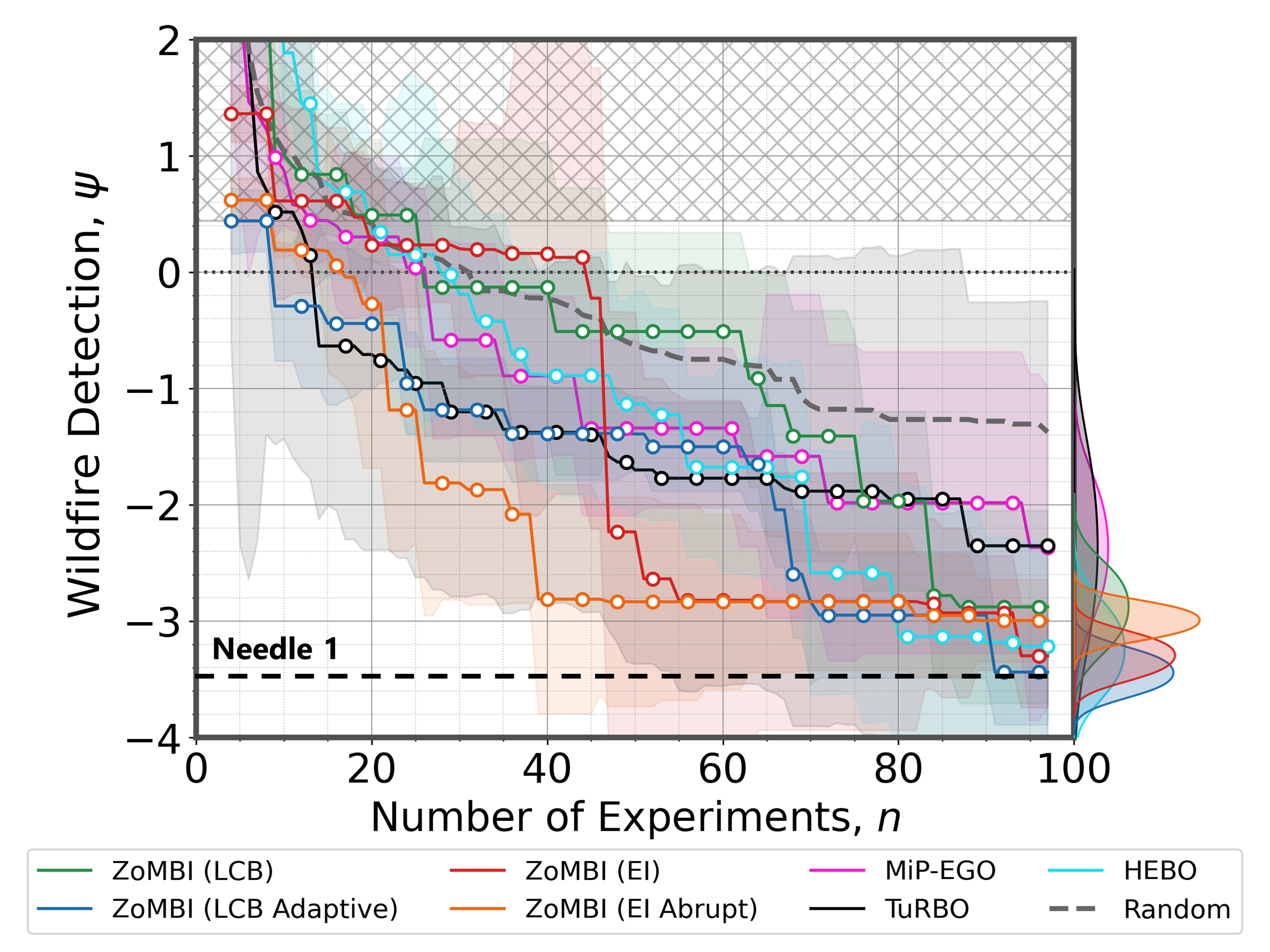}
\end{subfigure}\hfill%
\caption{\textbf{Detection of Environmental Conditions with Wildfire Risk.} The optimization objective is to find the meteorological conditions with the minimum wildfire detection index, $\psi$, in $100$ experiments from the dataset presented in Figure \ref{fig:all_hists}(c). Conditions with $\psi<0$ have the highest risk of sustaining wildfire. The green, blue, red, and orange lines indicate the median best running evaluated sample of \texttt{ZoMBI} using the LCB, LCB Adaptive, EI, and EI Abrupt acquisition functions, respectively. The pink, black, and teal lines indicate the median best running evaluated sample of the methods \texttt{MiP-EGO}, \texttt{TuRBO}, and \texttt{HEBO} respectively. Random sampling is illustrated as a dashed grey line for benchmarking. The median for each method is taken over the best $12$ independent model runs. The shaded regions indicate the variance between model runs. The cross-hatched region indicates the space discovered by standard BO methods, without the use of \texttt{ZoMBI}, which use the same hyperparameters as described in Section \ref{sec:acquisition}. The distribution across all $12$ model runs of the final sampled experiment for each method is shown as a kernel density estimation (KDE) along the $y$-axis. The y-values for the needle-like optima are indicated by dashed black lines.}
\label{fig:fire_bench}
\end{figure}

The third experimental dataset is 11-dimensional and consists of $128$k meteorological conditions and an index, $\psi$, that determines whether the set of conditions has a high likelihood of generating or sustaining a wildfire in the state of California -- publicly available from the California Irrigation Management Information System (CIMIS) weather stations \cite{Kohli2020}. Only $4.16\%$ of the total $128$k meteorological conditions have a negative wildfire detection index, $\psi < 0$. A highly negative $\psi$ indicates a high risk of wildfires. Hence, for this experiment, we aim to \textbf{minimize} $\mathbf{\psi}$, to best detect meteorological conditions at high risk of wildfires. The dataset spans over two years of data collected from 2018 to 2020, during which over $2500$ wildfires have occurred, burning over 24 million acres of land \cite{Mohapatra2022}. In California, temperature and precipitation alone are poor indicators for wildfire outbreaks (see Figure \ref{fig:shap}(c)), resulting in researchers using computer-vision methods or convolutions of many meteorological variables to reliably detect wildfire conditions instead \cite{Bouguettaya2022, Mohapatra2022}. Thus, there is a high need for algorithmic support to aid humans in early wildfire detection.

Figure \ref{fig:fire_bench} demonstrates the optimization performance of \texttt{ZoMBI} on the Wildfire Detection dataset compared to \texttt{MiP-EGO}, \texttt{TuRBO}, and \texttt{HEBO}. In this experiment, LCB Adaptive, EI, and \texttt{HEBO} discover the lowest index value, $\psi \approx -3.5$, for detecting wildfires based on a high-dimensional convolution of ten meteorological variables. \texttt{TuRBO} and \texttt{MiP-EGO} also discover a low index value, $\psi \approx -2.5$, however, these methods have widely distributed variances, as shown by the KDE plots, indicating inconsistent optimization results given only 100 sampled experiments. Similarly, \texttt{HEBO} has high variance across model runs while the LCB Adaptive and EI \texttt{ZoMBI} methods have a tight distribution, indicating more reliable optimization results with a higher rate of success. Furthermore, \texttt{ZoMBI} methods achieve a faster rate of convergence than \texttt{HEBO} onto the Needle 1 optimum, similar to the optimization results on the Poisson's Ratio dataset in Section \ref{sec:poisson}.

Figure \ref{fig:all_hists}(c) illustrates the distribution of $\psi$ values within the full dataset. The ground truth "needle" conditions for detecting wildfires are those with the most negative detection index values, $\psi$. Although \texttt{ZoMBI} with the LCB Adaptive and EI acquisition functions as well as \texttt{HEBO} discover the lowest needle-like $\psi$ conditions after 100 sampled experiments, none of the tested methods are able to find the global $\psi_\textrm{min}\approx -12$. These results imply that, even for \texttt{ZoMBI}, with a narrow enough needle-like optimum, the LHS initialization of $i=5$ experiments, may not be sufficient. In the next section, we stress-test this hypothesis on \texttt{ZoMBI}.

\section{Summary \& Conclusions}

In this paper, we proposed the [Zo]oming [M]emory-[B]ased [I]nitialization (\texttt{ZoMBI}) algorithm that builds on the principles of Bayesian optimization to accelerate the optimization of Needle-in-a-Haystack problems by two-fold, firstly by requiring fewer experiments to achieve better optimum faster than existing \texttt{MiP-EGO} \cite{van2019automatic}, \texttt{TuRBO} \cite{Eriksson2020}, and \texttt{HEBO} \cite{hebo} on a variety of real-world applications, and secondly by pruning the memory of low-performing historical experiments to speed-up compute time. The \texttt{ZoMBI} algorithm convergences onto narrow and sharp optima quickly in Needle-in-a-Haystack datasets by (1) using the values of the $m$ best performing previously sampled memory points to iteratively zoom in the search bounds of the manifold uniquely on each dimension and (2) implementing two custom acquisition functions, LCB Adaptive and EI Abrupt, that adapt their hyperparameters to tune sampling of new experimental conditions based on learned information from the surrogate model. The main contributions of this algorithm solve three fundamental challenges of optimizing non-convex Needle-in-a-Haystack problems: (1) the challenge of locating the hypervolume region of the manifold containing the narrow global optimum basin of attraction \cite{Crammer2004,Liu2018,Andricioaei1996} is alleviated by introducing iterative search bounds based on learned knowledge of the manifold; (2) the challenge of polynomially increasing compute times of BO using a GP surrogate \cite{BelyaevMikhail2014, Li2017, Wang2017, snelson2005, Rasmussen2005, Bui2017, Lan2020} is addressed by actively pruning the retained memory of the algorithm after each activation, $\alpha$, in turn, reducing the time complexity from $O(n^3)$ to $O(\phi^3)$ for $\phi$ forward experiments per activation, $\alpha$, which trends to a constant $O(1)$ when $\alpha>1$; (3) unwanted pigeonholing into local minima \cite{Nusse1996, Datseris2022, snelson2005, Rasmussen2005} is avoided by both the zooming mechanics of \texttt{ZoMBI} as well as using the two acquisition functions developed in this paper, LCB Adaptive and EI Abrupt, that tune their hyperparameters through adaptive learning. By developing the \texttt{ZoMBI} algorithm to solve these challenges, it becomes possible to quickly and efficiently find optimal solutions to complex Needle-in-a-Haystack problems in fewer experiments. 

Solving a Needle-in-a-Haystack problem that arises from extremely imbalanced data is a significant challenge that has important implications in science and engineering, especially within the field of materials science \cite{Kim2020, Crammer2004}. In this paper, we use \texttt{ZoMBI} to discover the optimum materials in two real-world materials science Needle-in-a-Haystack datasets where only a small fraction of the entire search space consists of the target optimum conditions. For breadth, we also extend our analysis to a third real-world dataset but for ecological resource management with the objective of discovering the environmental conditions that have a high likelihood of sustaining wildfires for early detection of wildfires. In the first materials dataset, we discover a material with a highly negative Poisson's ratio, $\nu$, \cite{Jain2013, DeJong2015}; in the second materials dataset, we discover a material with a highly positive thermoelectric figure of merit, $ZT$ \cite{Madsen2006, Jain2013}, both rare material properties; and in the third dataset for ecological resource management, we discover a set of environmental conditions with a highly negative wildfire detection index, $\psi$ \cite{Kohli2020, Bouguettaya2022, Mohapatra2022}. For the first dataset, both the \texttt{ZoMBI} algorithm with the LCB and LCB Adaptive custom acquisition functions and \texttt{HEBO} \cite{hebo} discover the material with the minimum $\nu \approx -1.7$, however, the \texttt{ZoMBI} methods converge on this minimum in only 70 experiments while \texttt{HEBO} takes 90 experiments. \texttt{TuRBO} \cite{Eriksson2020} and \texttt{MiP-EGO} \cite{van2019automatic} only discover materials with $\nu \approx -0.55$ and $\nu \approx -0.20$, respectively. For the second dataset, the \texttt{ZoMBI} algorithm with the LCB Adaptive custom acquisition function discovers the material with the maximum $ZT \approx 1.4$, while \texttt{HEBO} \cite{hebo}, \texttt{TuRBO} \cite{Eriksson2020}, and \texttt{MiP-EGO} \cite{van2019automatic} only discover $ZT \approx 0.78$, $ZT \approx 0.65$, and $ZT \approx 0.45$, respectively. For the third dataset the \texttt{ZoMBI} algorithm with all acquisition functions and \texttt{HEBO} \cite{hebo} discover a minimum $\psi \approx -3$, while \texttt{TuRBO} \cite{Eriksson2020} and \texttt{MiP-EGO} \cite{van2019automatic} both only discover $\psi \approx -2$. However, the \texttt{ZoMBI} methods converge on the minimum faster and with less variance. In general, we note \texttt{HEBO} \cite{hebo} outperforms the other benchmark methods, \texttt{TuRBO} \cite{Eriksson2020} and \texttt{MiP-EGO} \cite{van2019automatic}. Thus, for future investigation, we believe the performance of \texttt{ZoMBI} may be further improved by running optimization within the latent space of a variational autoencoder, similar to \texttt{HEBO} \cite{Maus2022, Grosnit2021}. Overall, these results demonstrate that the \texttt{ZoMBI} algorithm is more well-suited to tackle various real-world Needle-in-a-Haystack optimization problems than current methods, however, \texttt{ZoMBI} has performance limitations for extremely narrow optima when instantiated with an insufficient initialization set. Therefore to assess these limitations, we stress tested \texttt{ZoMBI} on an additional 174 analytical datasets with varying optimum needle widths, optimum distance to edges, dimensionality, and initialization conditions. These results concluded that with a fixed initialization set of 5 samples, \texttt{ZoMBI} has ideal performance on datasets with needle-like optima consisting of between $0.05\%$--$5\%$ of total hypervolume space. Furthermore, by extending the range of the initialization set, \texttt{ZoMBI} is capable of discovering global minima that lay on the absolute edge of a manifold's limits. Thus, in these certain cases, convergence to a global optimum using \texttt{ZoMBI} is not guaranteed, but with slight modifications based on some \textit{a priori} domain knowledge of the optimization landscape, \texttt{ZoMBI} produces high-performance and low-variance results.

Ultimately, the significance of developing the \texttt{ZoMBI} algorithm is to quickly and efficiently tackle difficult Needle-in-a-Haystack optimization problems in extremely imbalanced datasets. In this paper, we showcased the ability of the developed algorithm to discover rare materials and conditions with highly-optimized properties in a short period of time using few experiments. Discovering rare materials quickly and efficiently enables widespread access to a new range of materials applications from engineering high-performance medical devices to ubiquitous solid-state cooling systems  \cite{Saxena2016, Liu2006, Salah2020, He2018, Kim2020, Mao2020}. However, the application space for \texttt{ZoMBI} to accelerate the efficient discovery of highly-optimized solutions extends past materials science and is generally applicable for many Needle-in-a-Haystack problems, including those found in ecological resource management \cite{Rew2006, Bouguettaya2022}, fraud detection \cite{Wei2012,Marchant2021}, and rare disease prediction \cite{Khalilia2011, Marchant2021}. We aim for this contribution to support the elimination of the time and resource barriers previously inhibiting the throughput of optimizing complex and challenging Needle-in-a-Haystack problems across a broad range of application spaces.

\section*{Acknowledgements}
Basita Das is thanked for help in naming the algorithm. Xiaonan Wang is thanked for initial discussions for this study. John Dagdelen, Hongbin Zhang, and Shyam Dwaraknath are thanked for discussion of and reference to different Needle-in-a-Haystack problems within materials science. The authors acknowledge the MIT SuperCloud and Lincoln Laboratory Supercomputing Center for providing HPC resources that have contributed to the research results reported within this paper. This material is based upon the work supported by the U.S. Department of Energy’s Office of Energy Efficiency and Renewable Energy (EERE) under the Solar Energy Technology Office (SETO) award number DE-EE0009366.

\section*{Data Availability}
Implementation of the \texttt{ZoMBI} algorithm, the experimental dataset analyzed during the current study, the simulated data and labeled data supporting the findings of this study, and the data comprising the figures in this paper are all available in the following GitHub repository: \href{https://github.com/PV-Lab/ZoMBI}{https://github.com/PV-Lab/ZoMBI}.

\section*{Author Contributions}
A.E.S., Z.R., and T.B. conceived of and designed the study. Q.L. and T.B. provided guidance on machine learning methods, benchmark functions, and datasets. A.E.S. and Z.R. wrote the code. A.E.S. performed the machine learning modeling and analysis. A.E.S. wrote the paper, while all co-authors reviewed the manuscript.

\section*{Conflicts of Interest}
Although our laboratory has IP filed covering photovoltaic technologies and materials informatics broadly, we do not envision a direct COI with this study, the content of which is open sourced. Two of the authors (Z.R. and T.B.) own equity in Xinterra Pte Ltd, which applies machine learning to accelerate novel materials development.

\typeout{}
\bibliographystyle{unsrt}  
\bibliography{references}

\newpage
\beginsupplement
\begin{centering}
\LARGE{\textbf{Supporting Information:\\Fast Bayesian Optimization of Needle-in-a-Haystack Problems using Zooming Memory-Based Initialization (ZoMBI)}} \par
\end{centering}

\begin{description}
\begin{centering}

\item Alexander E. Siemenn$^{*,\dagger}$, Zekun Ren$^{\ddagger}$, Qianxiao Li$^{\mathparagraph}$, Tonio Buonassisi$^{\dagger}$
\item ${\dagger}$Department of Mechanical Engineering, Massachusetts Institute of Technology, Cambridge, MA 02139, USA
\item ${\ddagger}$Department of Electrical and Computer Engineering, Singapore-MIT Alliance for Research and Technology, Singapore 138602, Singapore
\item ${\mathparagraph}$  Department of Mathematics, National University of Singapore, Singapore 138602, Singapore
\item $^*$Corresponding author: asiemenn@mit.edu

\end{centering}
\end{description}

\section{Ackley Function Description}
\label{sec:ackley}

The Ackley function is an analytical function used in this paper to simulate a Needle-in-a-Haystack problem and is capable of having dimensionality between $1$-dimension and $d$-dimensions. The global minimum or "the needle" is challenging to discover in few experiments because the percentage of space made up by the global \textit{basin of attraction} decreases exponentially with dimensionality. The basin of attraction is the region containing the global minimum that once discovered, can be simply descended to the discover the global minimum. As the $b$ value of the Ackley function increases, the basin of attraction region narrows. In addition to having this needle-like global minimum, the Ackley function also has many local minima, generating a non-convex response surface. This response surface of the Ackley function is quantified through the following equation \cite{ackley1987}: 
\begin{equation}
\label{eq:ackley}
    f(X) = -a\;\textrm{exp}\left(-b \sqrt{\frac{1}{d}\sum_{i=1}^d X^2_i}\right) - \textrm{exp}\left(\frac{1}{d}\sum_{i=1}^d\textrm{cos}(c X_i)\right) + a + \textrm{exp}(1),
\end{equation}
where $a$ determines the connectivity of local minima, $b$ determines the width of the global minimum basin of attraction, $c$ determines the periodicity of the local minima, and $d$ is the number of dimensions. For this paper, the Ackley functions used have minima connectivity of $a=20$ and high minima periodicity of $c=\pi$ with $b$ and $d$ varying for most applications, but generally $b=0.5$, $d=5$. The response variable, $f(X)\in \mathbb{R}:[0,25]$ with the global minimum at $f(X)=0$ and the basin of attraction existing between $0\leq f(X) < 10$.


\section{Dataset Variables and Descriptions}

\begin{figure}[h!]
\centering
\begin{subfigure}[b]{.6\textwidth}  
\includegraphics[width=\textwidth]{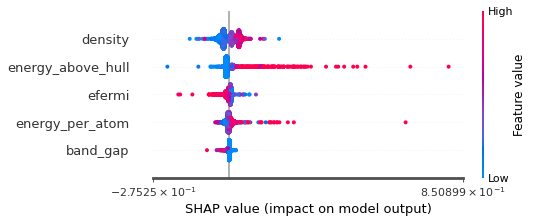}
\caption{Poisson's Ratio}
\end{subfigure}\hfill%
\begin{subfigure}[b]{.6\textwidth}  
\includegraphics[width=\textwidth]{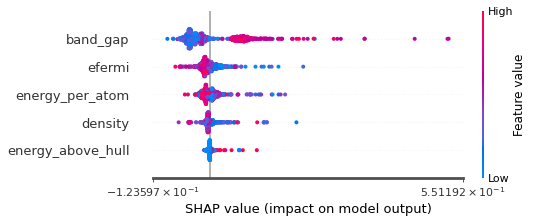}
\caption{Thermoelectric Merit}
\end{subfigure}\hfill%
\begin{subfigure}[b]{.6\textwidth}  
\includegraphics[width=\textwidth]{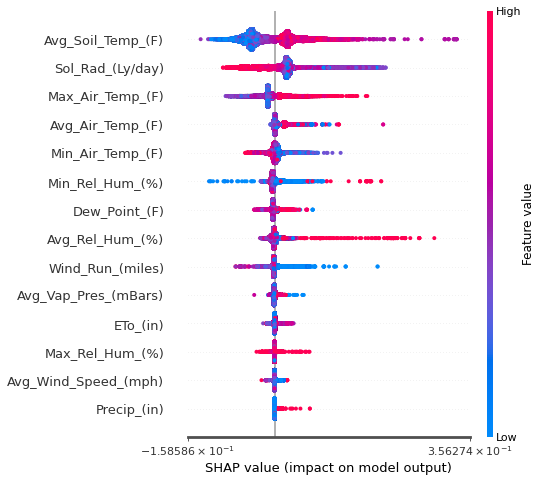}
\caption{Wildfire Detection}
\end{subfigure}\hfill%
\caption{\textbf{SHAP Feature Importance Ranking.} Each of the training variables for each real-world dataset have their importances ranked based on which variable has the highest impact on the output target variable. Variables with a higher spread of points have a higher impact on the target variable. Red scatter points are training variable values that increase the target variable output, while blue scatter points are training variable values that decrease the target variable output. The SHAP analysis was conducted by training a random forest regressor with an 80/20 training/testing split on the raw data points from each dataset.}
\label{fig:shap}
\end{figure}

In this paper, we provide analyses of three real-world datasets: (1) Poisson's Ratio \cite{Jain2013, DeJong2015}, (2) Thermoelectric Merit \cite{Madsen2006}, and (3) Wildfire Detection \cite{Kohli2020}. The Poisson's Ratio dataset has a sample size of $N=146,232$ experimental and simulated materials, publicly available from the Materials Project database \cite{Jain2013}. For this paper, we select five real-valued material variables to optimize Poisson's Ratio over, these variables are shown in Table \ref{table:poisson}. The feature importance ranking of these variables with the target variable is shown in Figure \ref{fig:shap}(a). The target variable, Poisson's Ratio is measured directly or computed using Density Functional Theory (DFT) from the stress, elasticity, and strain tensors, $\sigma_{ij}$, $C_{ijkl}$, and $\epsilon_{kl}$ \cite{DeJong2015}:

\begin{gather}
    \begin{bmatrix}
        \sigma_1 \\
        \sigma_2 \\
        \sigma_3 \\
        \sigma_4 \\
        \sigma_5 \\
        \sigma_6
    \end{bmatrix} 
 =
    \begin{bmatrix}
    C_{11} & C_{12} & C_{13} & C_{14} & C_{15} & C_{16} \\
    C_{12} & C_{22} & C_{23} & C_{24} & C_{25} & C_{26} \\
    C_{13} & C_{23} & C_{33} & C_{34} & C_{35} & C_{36} \\
    C_{14} & C_{24} & C_{34} & C_{44} & C_{45} & C_{46} \\
    C_{15} & C_{25} & C_{35} & C_{45} & C_{55} & C_{56} \\
    C_{16} & C_{26} & C_{36} & C_{46} & C_{56} & C_{66} 
    \end{bmatrix} 
    \begin{bmatrix}
        \epsilon_1\\
        \epsilon_2\\
        \epsilon_3\\
        2\epsilon_4\\
        2\epsilon_5\\
        2\epsilon_6
    \end{bmatrix} 
\end{gather}

The Thermoelectric Merit dataset has a sample size of $N=1,063$ experimental and simulated materials, publicly available from the Materials Project database \cite{Jain2013}. For this paper, we select five real-valued material variables to optimize Thermoelectric Merit over, these variables are shown in Table \ref{table:zt}. The feature importance ranking of these variables with the target variable is shown in Figure \ref{fig:shap}(b). The target variable, Thermoelectric Merit is computed from a combination of several variables as described in Equation \ref{eq:zt} using the BoltztraP software \cite{Madsen2006}. The Wildfire Detection dataset has a sample size of $N=128,009$ meteorological conditions measured on a daily basis across several weather stations in California in the CIMIS network \cite{Kohli2020}. For this paper, we select ten real-valued meteorological variables to optimize the Wildfire Detection Index over, these variables are shown in Table \ref{table:fire}. The feature importance ranking of these variables with the target variable is shown in Figure \ref{fig:shap}(c). The target variable, Wildfire Detection Index is obtained from \href{https://www.openml.org/search?type=data&status=active&id=43606}{the OpenML database}.

\begin{table}[h!]
\caption{Poisson's ratio dataset variables and description.}
\begin{tabular}{lll}
\hline
\textbf{Training Variable}   & \textbf{Units}               & \textbf{Description}                                                                    \\ \hline
Density           & g/cm$^3$ & Density of the entire molecule.                                                                            \\
Formation Energy  & eV/atom             & Normalized change of energy to form target phase.\\
Energy Above Hull & eV/atom             & Normalized energy to decompose into stable phase.\\
Fermi Energy      & eV                  & Highest energy level at absolute zero.\\
Band Gap          & eV                  & Valence to conduction band electron excitation energy. \\
                  &                     &                                                                                                            \\ \hline
\textbf{Target Variable}   & \textbf{Units}               & \textbf{Description}                                                                                                \\ \hline
Poisson's Ratio, $\nu$   & Unitless            & Mechanical deformation perpendicular to the loading direction.
\label{table:poisson}
\end{tabular}
\end{table}

\begin{table}[h!]
\caption{Thermoelectric Merit dataset variables and description.}
\begin{tabular}{lll}
\hline
\textbf{Training Variable}   & \textbf{Units}               & \textbf{Description}                                                                    \\ \hline
Density           & g/cm$^3$ & Density of the entire molecule.                                                                            \\
Formation Energy  & eV/atom             & Normalized change of energy to form target phase.\\
Energy Above Hull & eV/atom             & Normalized energy to decompose into stable phase.\\
Fermi Energy      & eV                  & Highest energy level at absolute zero.\\
Band Gap          & eV                  & Valence to conduction band electron excitation energy. \\
                  &                     &                                                                                                            \\ \hline
\textbf{Target Variable}   & \textbf{Units}               & \textbf{Description}                                                                                                \\ \hline
Thermoelectric Merit, $ZT$   & Unitless            & Electrical and thermal potential of a material to produce current.
\label{table:zt}
\end{tabular}
\end{table}

\begin{table}[h!]
\caption{Wildfire detection dataset variables and description.}
\begin{tabular}{lll}
\hline
\textbf{Training Variable}   & \textbf{Units}               & \textbf{Description}                                                                    \\ \hline
Average Soil Temperature           & $^\circ$F & Daily average measurement of the warmth in the soil.                                                                            \\
Solar Radiation  & Ly/day             & Amount of solar electromagnetic radiation received.\\
Max Air Temperature & $^\circ$F             & Daily maximum measurement of warmth in the air.\\
Dew Point      & $^\circ$F                  & Daily air temperature at which water condenses.\\
Min Air Temperature          & $^\circ$F                  & Daily minimum measurement of warmth in the air. \\
Min Relative Humidity           & \% & Daily minimum percentage of water vapor in air from saturation point.                                                                            \\
Average Vapor Pressure  & mbar             & Daily average partial pressure of water vapor in air.\\
Average Wind Speed & mph             & Daily average velocity of wind.\\
Evapotranspiration      & in                  & Amount of water lost from soil and plants to the atmosphere.\\
Precipitation          & in                  & Amount of rainwater received. \\
                  &                     &                                                                                                            \\ \hline
\textbf{Target Variable}   & \textbf{Units}               & \textbf{Description}                                                                                                \\ \hline
Wildfire Detection Index, $\psi$   & Unitless            & Metric for determining an environment's ability to sustain a wildfire.
\label{table:fire}
\end{tabular}
\end{table}

\newpage
\clearpage 
\section{Cumulative Compute Time}

\begin{figure}[h!]
\centering
\begin{subfigure}[b]{0.6\textwidth}  
\includegraphics[width=\textwidth]{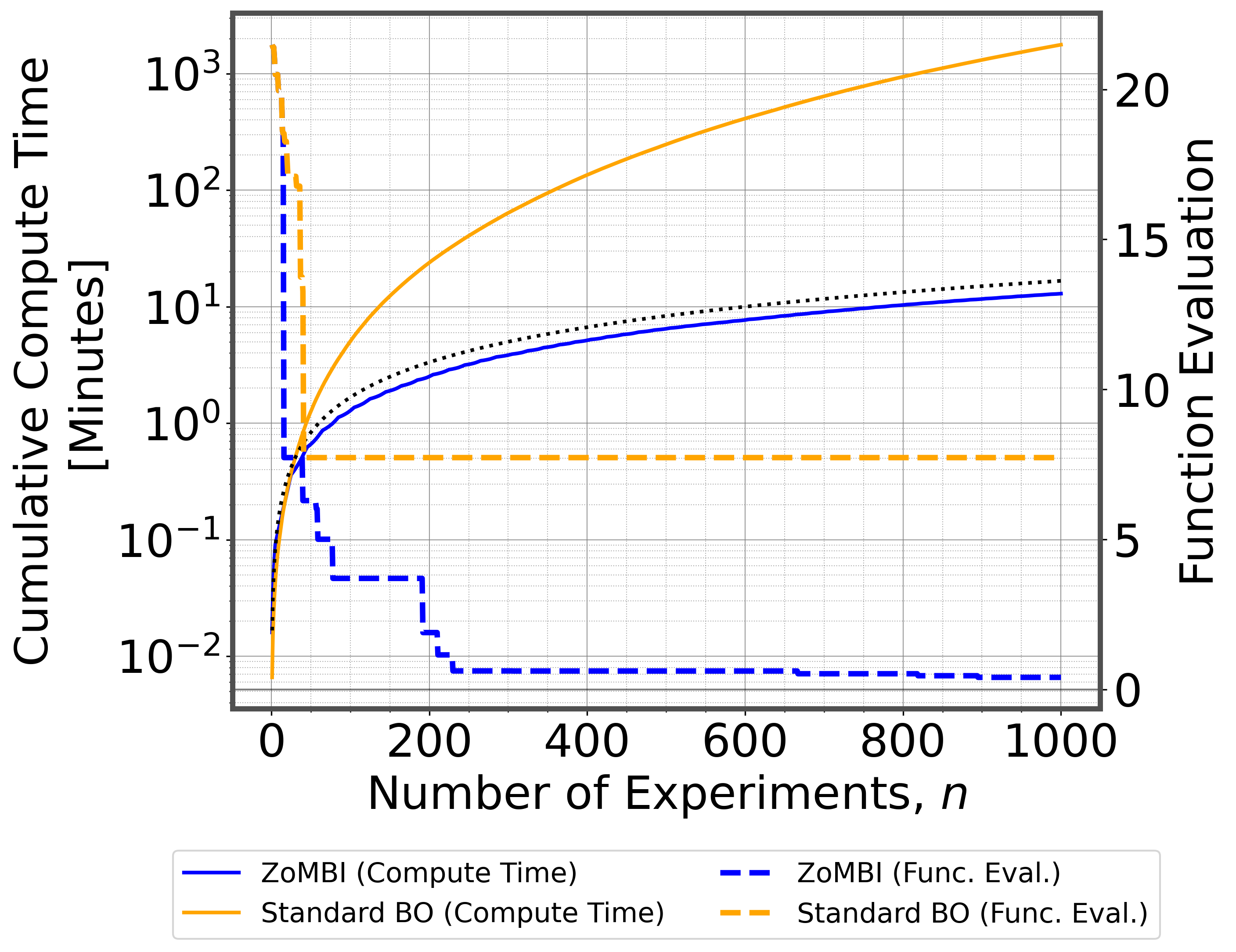}
\end{subfigure}\hfill%
\caption{\textbf{Cumulative Wall-clock Compute Time and Function Evaluations.} The median total cumulative compute time to sample 1,000 experiments of a 6D Ackley function is compared for four independent trials of standard BO (solid yellow line) and \texttt{ZoMBI} (solid blue line). Wall-clock compute time is measured on an NVIDIA Tesla Volta V100 GPU \cite{reuther2018interactive}. The black dotted line indicates the cumulative sum of a constant 1 second compute per experiment -- the compute time of \texttt{ZoMBI} lies below this line. The yellow and blue dashed lines correspond to the minimum evaluated point of the Ackley function for standard BO and \texttt{ZoMBI}, respectively, where zero is the global minimum. The cumulative compute times are plotted on a log-scale, whereas the function evaluations are plotted on a linear scale.}
\label{fig:cumcompute}
\end{figure}

In addition to showing the instantaneous compute times for both standard BO and \texttt{ZoMBI} in Figure \ref{fig:compute}, we show the cumulative compute times and the function evaluations achieved on the 6D Ackley function in Figure \ref{fig:cumcompute}. Each trace shows the average compute across four trials for both standard BO and \texttt{ZoMBI}, with each trial using one of the four acquisition functions: LCB, LCB Adaptive, EI, and EI Abrupt. The cumulative compute times give a sense of how long each algorithm took to complete the full optimization procedure of $1000$ experiments. To complete $1000$ experiments, standard BO took over 1000 minutes, whereas \texttt{ZoMBI} took only 10 minutes, two orders of magnitude shorter, with $\phi=10$ per activation. The black dotted line indicates the cumulative sum of a constant 1-second compute per experiment, we note that the compute time for \texttt{ZoMBI} is beneath that line, indicating that, on average, \texttt{ZoMBI} takes less than 1 second per experiment to compute. Furthermore, the minimum function evaluations are shown for both standard BO and \texttt{ZoMBI}. We show that with standard BO, the function evaluations plateau very early on, however, with \texttt{ZoMBI}, the function evaluations continue to decrease until the global minimum is discovered at around 900 experiments.

\section{Varying Basin of Attraction Width}

\begin{figure}[h!]
\centering
\begin{subfigure}[b]{0.8\textwidth}  
\includegraphics[width=\textwidth]{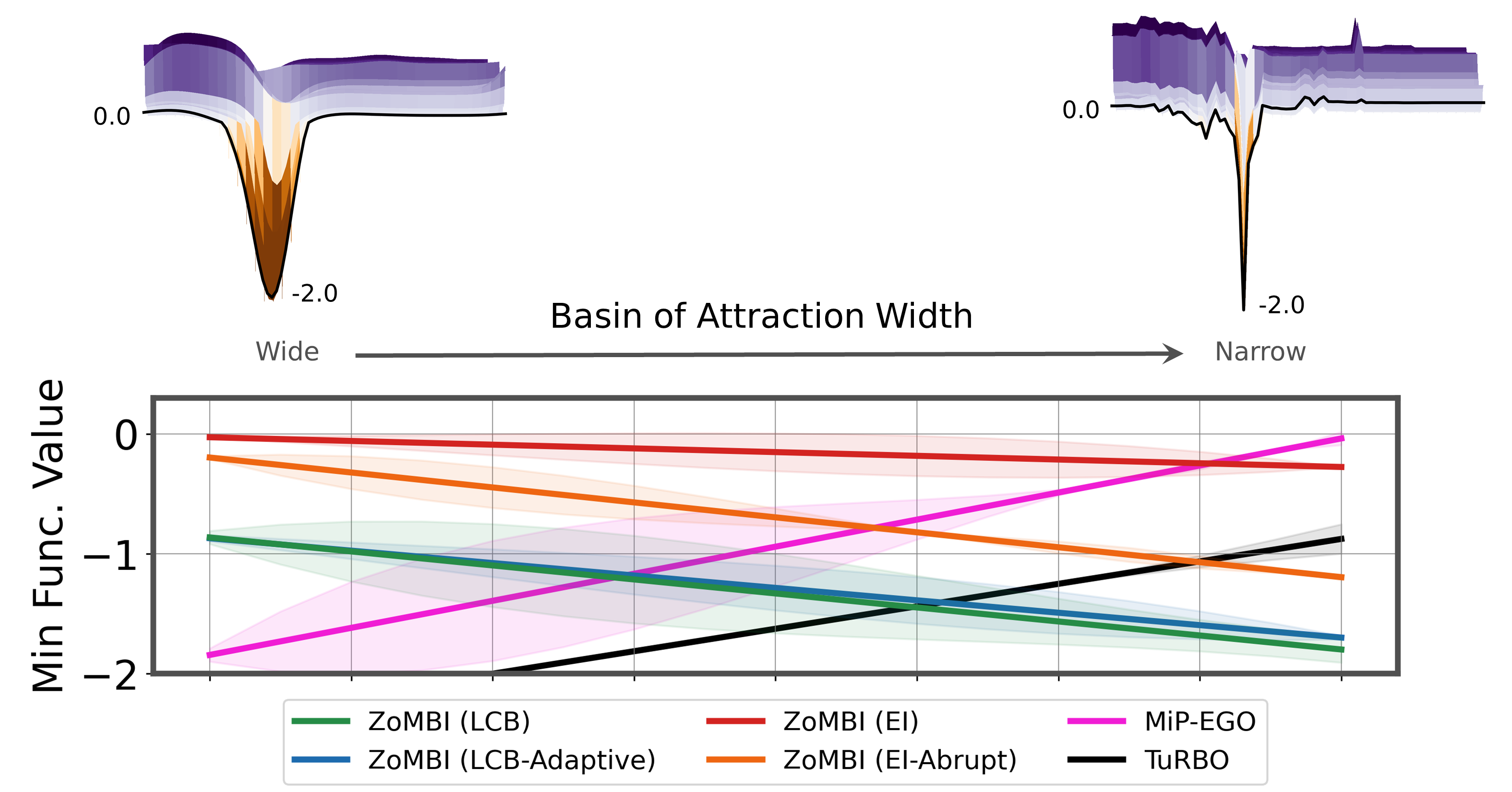}
\end{subfigure}\hfill%
\caption{\textbf{Synthetically Varying Basin of Attraction Width.} The colored lines indicate the median trend lines of $12$ independent runs for the final experimental evaluation of each algorithm after $100$ sampled experiments for a given basin width. The spread of the data from the trend line is indicated by the shaded regions. The datasets with different width basins are obtained by synthetically convolving the original 6D Poisson's ratio dataset topology with Gaussian noise of increasing magnitudes. The objective is to find the minimum function value, $f_\textrm{min}=-2$. In wide basin datasets, \texttt{TuRBO} and \texttt{MiP-EGO}, perform the best. However, for NiaH problems with narrow basins, the \texttt{ZoMBI} implementations perform better.}
\label{fig:smooth}
\end{figure}

The \texttt{ZoMBI} algorithm is designed specifically to tackle NiaH problems where the basin of attraction containing a global minimum is narrow \cite{Nusse1996, Datseris2022}. In this experiment, we explore the question, \textit{how does basin of attraction width affect the ability of \texttt{ZoMBI} to find the global minimum?} We present a synthetic example of a 6D noisy and non-convex dataset with a global minimum of $f_\textrm{min} = -2$ whereby applying a Gaussian smoothing function to the dataset, the basin of attraction width is synthetically widened. The original, unmodified 6D manifold is based on the Poisson's Ratio dataset \cite{Jain2013, DeJong2015}. Figure \ref{fig:smooth} illustrates the trend lines of optimization performance of \texttt{ZoMBI} with each of its possible acquisition functions, relative to \texttt{TuRBO} \cite{Eriksson2020} and \texttt{MiP-EGO} \cite{van2019automatic}. The trend lines show that for wider basin of attraction problems, \texttt{TuRBO} and \texttt{MiP-EGO} outperform \texttt{ZoMBI} with its various acquisition functions. However, as the basin of attraction narrows, the problem becomes a NiaH. The trend in performance of \texttt{ZoMBI} improves as the basin width decreases, the algorithm becomes more capable of finding the global minimum. Conversely, \texttt{TuRBO} and \texttt{MiP-EGO} become less capable of discovering the global minimum as the manifold transitions into a NiaH problem. As the Gaussian smoothing is lessened, all global and local minima basins narrow, resulting in less of the manifold space containing optimal regions, which accelerates the inward zooming of the search bounds using \texttt{ZoMBI}. Additionally, the more explorative acquisition functions are shown to perform better than the exploitative functions, such as EI, because pigeonholing is more readily avoided, as shown previously in Figure \ref{fig:search}.

\section{Optima Near Edges}
\label{sec:edge}

\begin{figure}[h!]
\centering
\begin{subfigure}[b]{0.8\textwidth}  
\includegraphics[width=\textwidth]{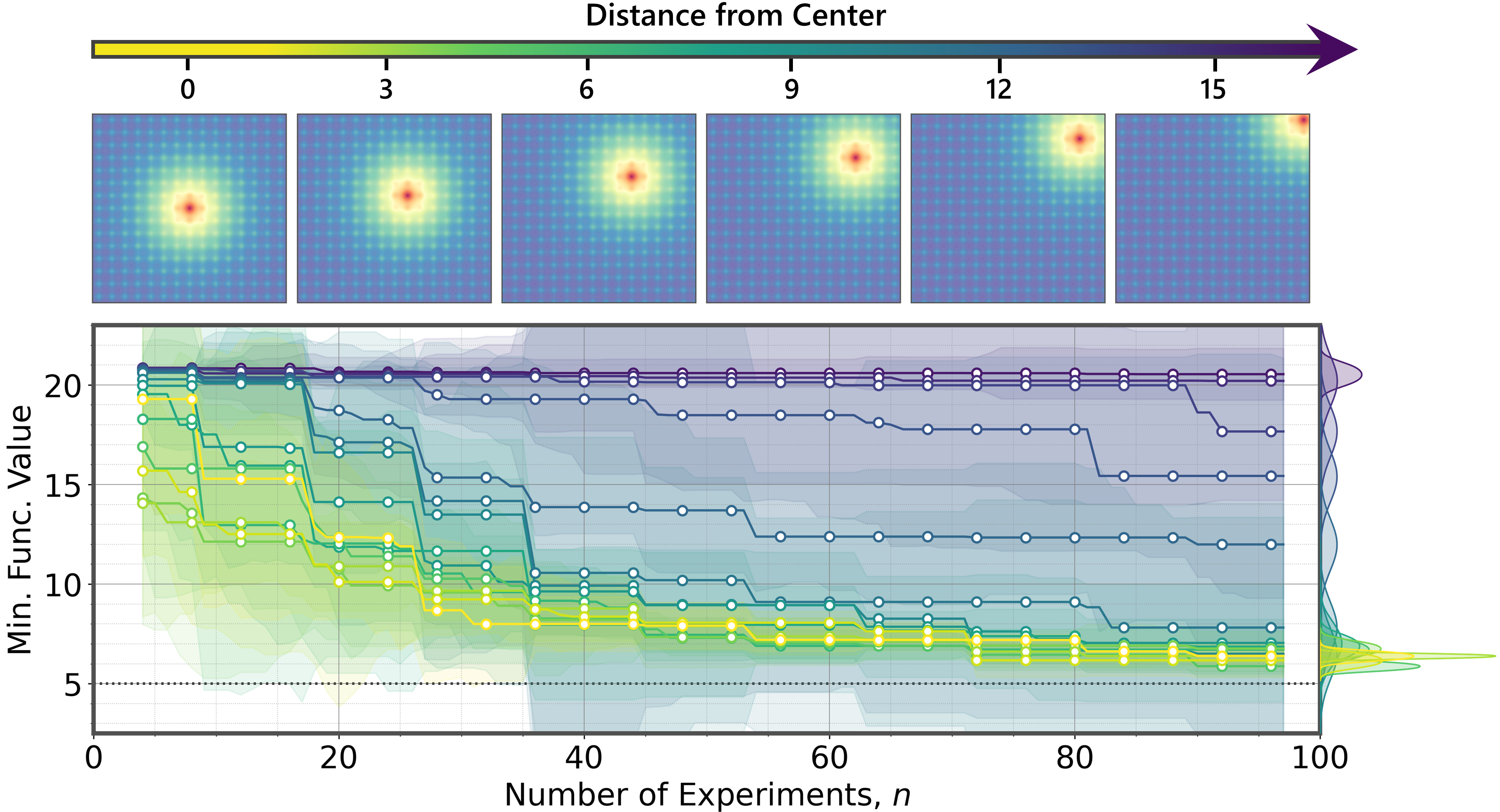}
\caption{Optima Distance from Center}
\vspace{1.5em}
\end{subfigure}\hfill%
\begin{subfigure}[b]{0.8\textwidth}  
\includegraphics[width=\textwidth]{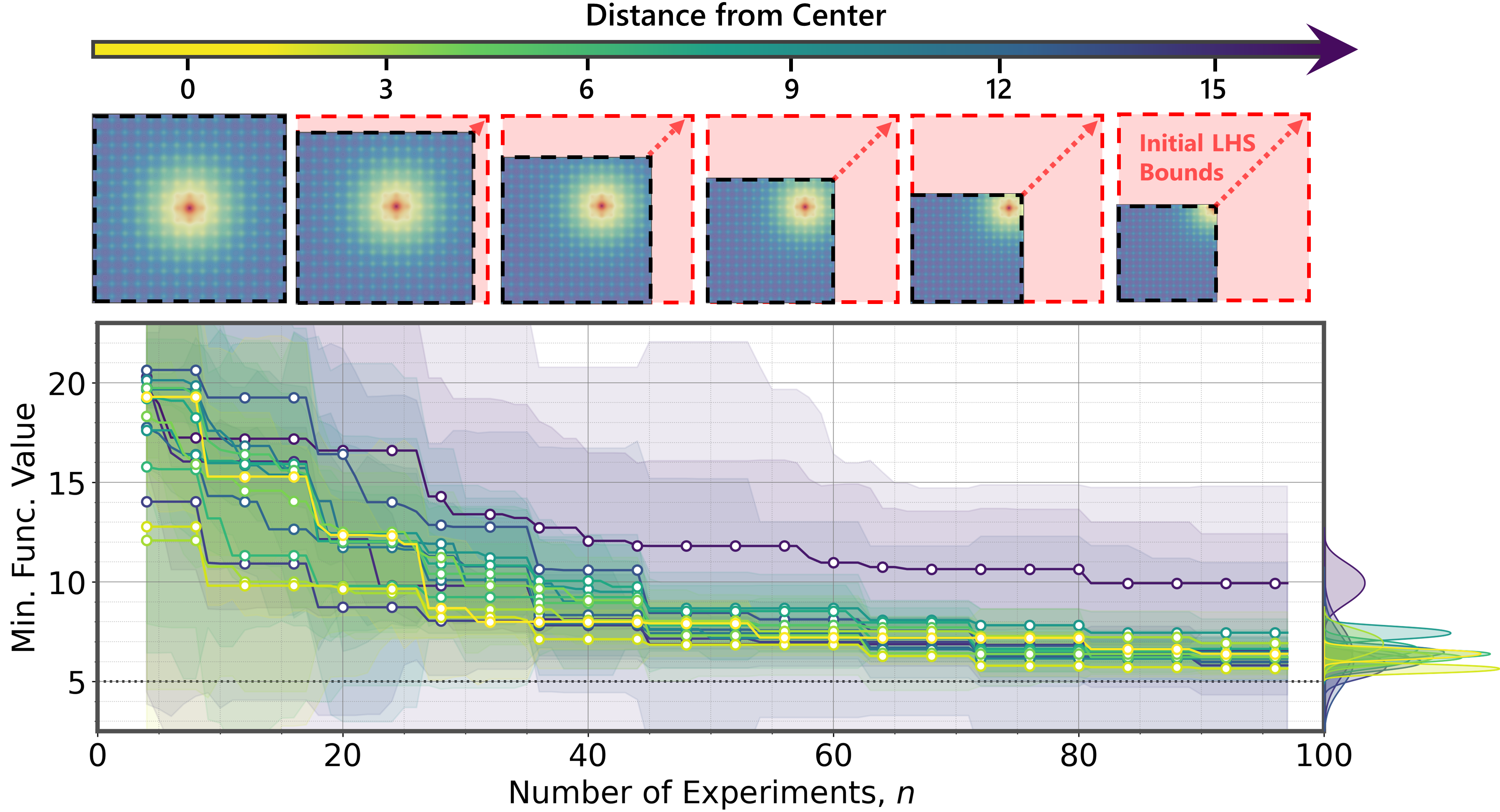}
\caption{Optima Distance from Center with Extended Initialization}
\end{subfigure}\hfill%
\caption{\textbf{Varying Optimum Edge Distance and Initialization Conditions.} (a) The performance of \texttt{ZoMBI} is shown to degrade if the global optimum needle of a manifold exists near a boundary with a constant initialization set. (b) However, by simply extending the initialization bounds for LHS sampling, this unfavorable degradation in performance is avoided. Each trace represents the median and variance of four independent trials of \texttt{ZoMBI} with the LCB Adaptive acquisition function over 100 evaluated experiments of a 6D Ackley function with hundreds of trap local minima. The colorbar of the traces represents the optimum distance from the center of the manifold. A distance of 15 represents an optimum at the upper limit of the hypervolume corner. The final variance across all trials is illustrated as a KDE plot along the $y$-axis.}
\label{fig:distance}
\end{figure}

A dataset with a global minimum near an edge is a challenge for \texttt{ZoMBI} to optimize unless the initialization set is altered. In this experiment, we evaluate the performance of \texttt{ZoMBI} across fifteen datasets, where each dataset moves the global optimum closer to an edge region of the manifold. All fifteen experiments are repeated but with an initialization set extending beyond the limiting edge of the manifold. Figure \ref{fig:distance}(a) illustrates the minimum function values discovered by \texttt{ZoMBI} over 100 experiments for each of the fifteen 6D datasets with varying optimum distances from the center, using an initialization set of $i=5$ with bounds $[0,1]^6$ for each run. It is shown that once the optima begin to approach the corner (between distance 12-15 from the center), \texttt{ZoMBI} is no longer able to discover the minimum and plateaus performance across 100 sampled experiments. These results are explained by the intertwined nature of the iterative zooming feature of \texttt{ZoMBI} and the initialization set used to instantiate the optimization procedure. If the initialization is at the edge of a small initialization set, an edge-bounded optimum may be easily excluded from the internal initialization hypervolume. Hence, when \texttt{ZoMBI} computes a GP surrogate and zooms in its bounds, the global minimum is excluded because it was not contained sufficiently within the initialization bounds. To surmount this unfavorable mechanic, one may either increase the number of initialization points or extend the bounds of LHS sampling if known \textit{a priori} that the global minimum of the manifold lies close to a boundary.

Figure \ref{fig:distance}(b) demonstrates that by extending the LHS sampling bounds during only the initialization of \texttt{ZoMBI}, performance degradation of edge-bounded optima is avoided entirely. Extending the initialization bounds allows the GP surrogate to extrapolate edge feature information, which \texttt{ZoMBI} can more reliably zoom inward on than features at the edge of the bounds. Since it was known \textit{a priori} that the global minimum of the tested datasets existed at the upper limit of the manifold hypervolume, the upper limit of the LHS initialization set was extended. For an arbitrary and normalized $d$-dimensional manifold with an optimum existing at the upper limit of the manifold, LHS bounds are simply extended from $[0,1]^d$ to $[0,1+\gamma]^d$, where $\gamma < 0$. Although \texttt{ZoMBI} does not work out-of-the-box for every application, simple modifications can be made to a single step within the procedure to significantly improve optimization results for a breadth of dataset topologies without increasing the required number of sampled experiments.

\end{document}